\documentclass[10pt,twocolumn,letterpaper]{article}

\usepackage{iccv}
\usepackage{times}
\usepackage{epsfig}
\usepackage{graphicx}
\usepackage{amsmath}
\usepackage{amssymb}
\usepackage{multirow}
\usepackage{diagbox}
\usepackage{subfig}
\usepackage{floatrow} 
\usepackage[font=small]{caption}

\newcommand{\inerf}{Instance-NeRF\xspace}
\newcommand{\inerflong}{Instance Neural Radiance Field\xspace}
\newcommand{\nerf}{NeRF\xspace}
\newcommand\width{-0.00cm}
\newcommand\height{-0.29cm}

\usepackage[breaklinks=true,bookmarks=false]{hyperref}

\iccvfinalcopy 

\def\iccvPaperID{10630} 

\ificcvfinal\fi

\begin{document}

\title{\inerflong}

\author{
\begin{tabular}{ccccc}
Yichen Liu$^{1}${\footnotemark[1]} & 
Benran Hu$^{2}$\thanks{ Equal contribution. } & 
Junkai Huang$^{2}${\footnotemark[1]} & 
Yu-Wing Tai$^{3}$ &
Chi-Keung Tang$^{1}$ 
\end{tabular}
\vspace{0.05in}
\\
\begin{tabular}{ccc}
$^1$The Hong Kong University of Science and Technology
\end{tabular}
\\
\begin{tabular}{ccc}
$^2$Carnegie Mellon University &
$^3$Dartmouth College
\end{tabular}
}

\maketitle
\ificcvfinal\thispagestyle{empty}\fi

\renewcommand{\thefootnote}{\fnsymbol{footnote}}
\footnotetext[2]{This research is supported in part by the Research Grant Council of the Hong Kong SAR under grant no. 16201420.}


\begin{abstract}
This paper presents one of the first learning-based NeRF 3D instance segmentation pipelines, dubbed as {\bf \inerflong}, or \inerf. Taking a NeRF pretrained from multi-view RGB images as input, \inerf can learn 3D instance segmentation of a given scene, represented as an instance field component of the NeRF model. To this end, we adopt a 3D proposal-based mask prediction network on the sampled volumetric features from NeRF, which generates discrete 3D instance masks. The coarse 3D mask prediction is then projected to image space to match 2D segmentation masks from different views generated by existing panoptic segmentation models, which are used to supervise the training of the instance field. Notably, beyond generating consistent 2D segmentation maps from novel views, \inerf can query instance information at any 3D point, which greatly enhances NeRF object segmentation and manipulation. Our method is also one of the first to achieve such results in pure inference. Experimented on synthetic and real-world NeRF datasets with complex indoor scenes, \inerf surpasses previous NeRF segmentation works and competitive 2D segmentation methods in segmentation performance on unseen views. Code and data are available at \url{https://github.com/lyclyc52/Instance_NeRF}.

\end{abstract}

\section{Introduction}
Neural Radiance Field (NeRF)~\cite{mildenhall2020nerf}  has become the mainstream approach to novel view synthesis nowadays. Given multi-view images with camera poses only, NeRF encodes the underlying scene in a multi-layer perceptron (MLP) by radiance propagation and generates very impressive results. Thus, subsequent to NeRF's debut in~\cite{mildenhall2020nerf} lot of works have made great progress in improving the quality~\cite{kangle2021dsnerf, wang2021nerf-sr}, efficiency~\cite{mueller2022instant, yu2021plenoctrees, Chen2022ECCV} and generality~\cite{zhang2020nerf++, wang2021nerf}.

\begin{figure}[t]
\centering
    
    \includegraphics[width=0.9\linewidth]{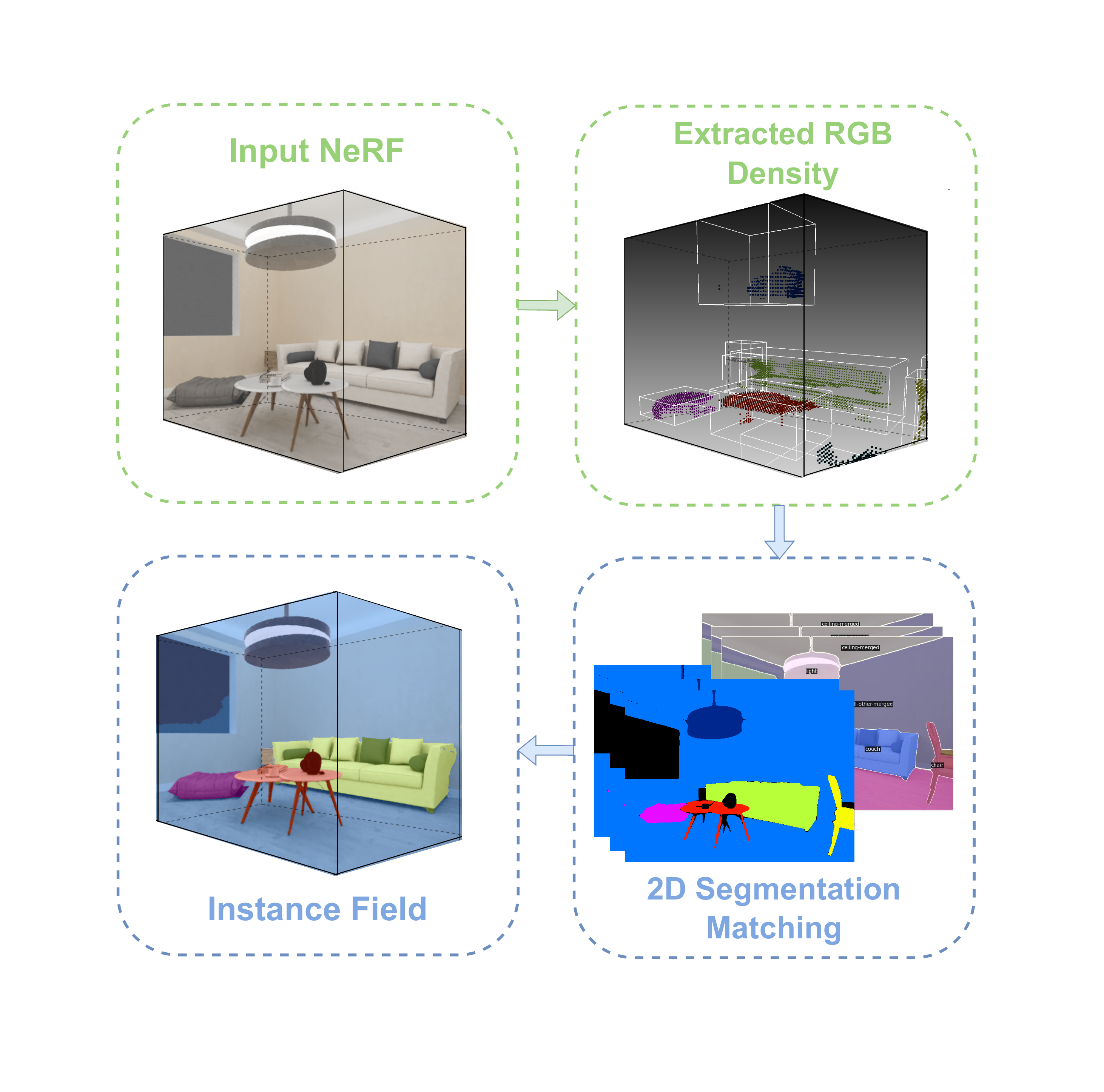}
    \vspace{0.15in}
    
    \caption{Pipeline of \inerf, demonstrated on the 3D-FRONT dataset. \inerf takes a pre-trained NeRF as input to detect objects within the scene and utilizes existing 2D panoptic segmentation to generate 2D segmentation maps, which are then matched and used to supervise the instance field training.}\vspace{-0.15in}
    \label{fig:teaser}
\end{figure}

This excellent approach to associate 2D with 3D through radiance field leads us to rethink the {\em 3D instance segmentation problem}. Unlike the 2D counterpart operating on plenty of training images, 3D instance segmentation is limited by both the quantity and quality of the available data. Previously, 
3D segmentation still relies on RGB-D images~\cite{hou2019sis} or point clouds~\cite{Qi_2017_CVPR, NIPS2017_d8bf84be, Wang_2018_CVPR, Wu_2015_CVPR, tchapmi2017segcloud} captured by depth sensors or custom equipment as input, which are inconvenient to obtain and contain a variety of noises. 
To mitigate the dependence on explicit 3D geometry, there have been several investigations on embedding NeRF for addressing fundamental problems such as 3D semantic segmentation~\cite{Zhi:etal:ICCV2021} and  scene manipulation~\cite{lazova2023control}. Existing unsupervised methods such as~\cite{liu2022unsupervised, yu2022unsupervised, stelzner2021decomposing} also involve 3D scene decomposition and instance segmentation, but it is hard to apply them on complex and large scenes akin to real world cases. 

In this paper, inspired by NeRF-RPN~\cite{nerf-rpn} and Mask-RCNN~\cite{He_2017_ICCV}, we propose {\em \inerf} for 3D instance segmentation in NeRF. Unlike semantic segmentation, where a single object in different views should be labeled as the same class consistently, most instance segmentation methods do not enforce consistency over instance IDs in different views, making direct supervision of instance ID less applicable. To address this issue, we propose a 2D mask matching procedure that links each 2D instance mask to its 3D instance to resolve  inconsistency. More specifically, we incorporate NeRF-RPN with a mask head to predict 3D coarse segmentation. After projecting the attained 3D masks back to 2D, \inerf leverages Mask2Former~\cite{cheng2021mask2former} and CascadePSP~\cite{cheng2020cascadepsp} to match the same instance in 2D segmentation from different views and refine the resultant masks. The refined 2D segmentation of multi-view images will be used to train an instance field which encode 3D instance information in a continuous manner as a neural field.

Our major contributions are:
\begin{itemize}
\setlength{\itemsep}{-2pt}
\item One of the first significant attempts to perform 3D instance segmentation in NeRF without using ground-truth segmentation as input.
\item Propose the architecture and training approach of an \textit{Neural Instance Field}, which can produce multi-view consistent 2D segmentation as well as continuous 3D segmentation using NeRF representation.
\item Perform experiments and ablation studies on a synthetic indoor NeRF dataset to demonstrate the effectiveness of our method, which surpasses competitive 2D segmentation methods and previous works in NeRF segmentation.
\end{itemize}

\section{Related Works}

\subsection{Neural Radiance Fields}
Neural Radiance Field (NeRF)~\cite{mildenhall2020nerf} is now the state-of-the-art for reconstructing new views of a given scene, by modeling the underlying 3D geometry and appearance using a continuous and implicit radiance field parameterized by a multilayer perceptron (MLP). Recent works include:
Instant Neural Graphics Primitive~\cite{mueller2022instant} uses hash encoding to speed up training, PlenOctrees~\cite{yu2021plenoctrees} uses an octree-based radiance field and spherical basis functions to improve rendering speed and appearance decoding, and TensoRF~\cite{Chen2022ECCV} encodes positional information by projecting 3D points onto three 2D planes. 
NeRF not only provides structural details of a 3D scene but is also conducive to 3D training, where only RGB images with camera parameters are required, thus making this alternative 
also suitable for 3D unsupervised object segmentation~\cite{liu2022unsupervised}, where first encouraging results are demonstrated on real scenes using radiance propagation but with no geometry consideration.

\subsection{R-CNN}
In 2D,  Region-based CNN (R-CNN)~\cite{Girshick_2014_CVPR} 
focuses on a small number of regions \cite{HosangBDS15}, and uses convolutional networks to independently analyze each RoI. R-CNN was extended by incorporating RoIPool \cite{HeZR014, Girshick_2015_ICCV} for attending to regions on feature maps, making it faster and more accurate. Faster R-CNN~\cite{renNIPS15fasterrcnn} incorporates Region Proposal Network (RPN) to learn the attention mechanism. Mask R-CNN~\cite{He_2017_ICCV} adds a branch to predict an object mask in parallel with the existing branch for bounding box recognition.

In 3D, NeRF-RPN~\cite{nerf-rpn} bridges RPN and NeRF and demonstrates great potential in direct 3D learning from NeRFs. This paper contributes to 3D object instance segmentation from NeRFs by capitalizing on the 3D boxes given by NeRF-RPN. In other words,  
analogous to Mask-RCNN, our work extends NeRF-RPN to enable 3D instance segmentation directly from a given NeRF. 

\subsection{Instance Segmentation}
\label{sec:inst_seg}
For 2D instance segmentation,
two-stage methods~\cite{li2017fully, He_2017_ICCV, cai2018cascade, chen2019hybrid, cheng2020boundary} 
first detect candidate bounding boxes and then predict instance mask within each of them. On the other hand,~\cite{chen2019tensormask, kuo2019shapemask, xie2020polarmask, bolya2019yolact, wang2020solo, wang2020solov2} release the network from proposal generation and achieve comparable results. Despite the great success of 2D instance segmentation, straightforward extension to 3D does not work in general. Specifically, 
applying 2D instance segmentation on each image capturing a single scene does not guarantee consistence across multi-view images of the underlying 3D scene.

Current methods on 3D instance segmentation usually perform on RGB-D images or point clouds. Image-based approaches such as~\cite{hou2019sis} use 2D convolution to extract RGB-D features and then project them back to infer 3D segmentation. For point-cloud-based methods, while some methods~\cite{Wu_2015_CVPR, tchapmi2017segcloud, Han_2020_CVPR, Vu_2022_CVPR, liang20203d} voxelize the inputs and adopt 3D convolution, others~\cite{Qi_2017_CVPR, NIPS2017_d8bf84be, Wang_2018_CVPR, yang2019learning} preserve the irregularity of point clouds and utilize permutation-invariant function to extract features. In addition, some approaches~\cite{ Dai_2018_ECCV, Jaritz_2019_ICCV} take RGB images as extra inputs and fuse features from two types of data. However, all of them require some explicit 3D geometry e.g., obtained by LiDARs or other devices. 3D instance segmentation directly from multi-view images has not been explored in the context of NeRFs.

Segmentation and decomposition on NeRF were investigated. Semantic-NeRF~\cite{Zhi:etal:ICCV2021} extends the NeRF by adding a semantic branch to produce view independent semantic labels. Semantic-NeRF predicts accurate semantic labels even when the ground truth labels are sparse and noisy, but this work mainly targets semantic map denoising and super-resolution instead of inference tasks. DM-NeRF~\cite{wang2022dmnerf}, on the other hand, uses multi-view ground-truth instance maps to optimize an object field for NeRF object decomposition and manipulation. Panoptic NeRF~\cite{fu2022panoptic} can render accurate semantic maps with sparse multi-view images and the corresponding noisy predicted masks given by a pre-trained model. However, it requires 3D bounding primitives to guide consistency and predict segmentation on instance level. Panoptic Neural Fields (PNF)~\cite{kundu2022panoptic} assumes objects are dynamic and tracks 3D object from images in panoptic images, which is not suitable for general scenes represented by NeRF. NeSF~\cite{vora2021nesf} utilizes a 3D UNet to generate semantic feature grid from density grid sampled from NeRF, and performs volume rendering over the feature grid to produce 2D semantic maps. NeSF does not need ground truth semantics during inference, but it cannot perform instance segmentation.

There have been some investigation on unsupervised object segmentation in NeRF. GIRAFFE~\cite{Niemeyer2020GIRAFFE} uses generative neural feature fields to decode shape and appearance, which decomposes foreground objects from background without explicit 3D segmentation learning.  ~\cite{stelzner2021decomposing, yu2022unsupervised} apply slot attention~\cite{locatello2020object} on unsupervised object discovery, while others~\cite{fan2022nerf, liu2022unsupervised} incorporate the advantages of NeRF to obtain multi-view 2D masks. However, these methods do not produce class labels for segmented objects, and only achieve qualified success in simple scenes with a limited number of objects. There is still large room for quality improvement and generalization.

3D instance segmentation in NeRF is a less explored and more challenging task. Unlike 2D semantic mask priors used in Semantic-NeRF, which are mostly consistent over different views, 2D instance masks are generally multi-view inconsistent, as the same object may be assigned different instance IDs in different views. This inconsistency can greatly impair the performance of Semantic-NeRF and DM-NeRF, leading to noisy semantic fields. While Panoptic NeRF partially solves this issue with 3D bounding volumes as priors, it cannot resolve the ambiguity when objects belonging to the same class have overlapping bounding volumes. In comparison, the \inerf we propose does not rely on 3D priors to address the inconsistency and avoid such ambiguity. Our work is one of the first learning-based approaches to generate 3D instance segmentation in NeRF, which can be applied to produce satisfactory 3D object masks on complex scenes 
during test time.

\section{Method}

\begin{figure*}[t]
\centering
    \includegraphics[width=1\linewidth]{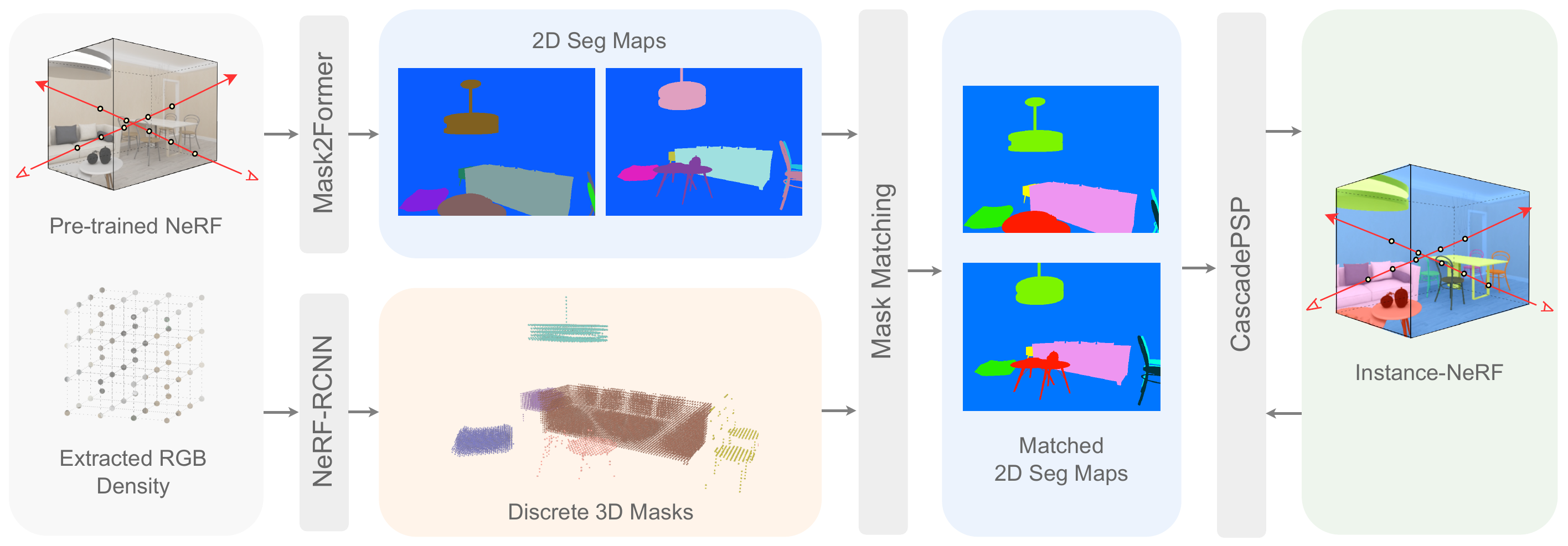}
    \vspace{-0.10in}
    
    \caption{\textbf{Instance Field Training.}  When training the instance field, NeRF-RCNN takes the extracted radiance and density field of the pre-trained NeRF and outputs discrete 3D masks for each detected object in the NeRF. Mask2Former generates 2D panoptic segmentation maps of images rendered from \nerf, which are \textit{inconsistent} in terms of instance label across views. After projecting the 3D masks from the same camera poses to match the same instance across different views resulting in the multi-view \textit{consistent} 2D segmentation maps,  they can then be used to train the instance field component and produce a continuous 3D segmentation using instance field representation. In addition, we use CascadePSP to refine the preliminary instance segmentation results of \inerf, and use the refined instance masks to refine the Instance Field in turn. }\vspace{-0.15in}
    \label{fig:main}
\end{figure*}

Given a pre-trained \nerf, our method, \inerf, aims to detect all the objects within the underlying 3D scene and produces a bounding box, a continuous 3D mask, and a class label of each detected 3D object. 

\inerf extends the given pre-trained \nerf model with an additional instance field, which can produce a view-independent instance label at any 3D position in the NeRF scene. 
To train the instance field component of \inerf, we propose a {\em NeRF-RCNN} for 3D sparse mask prediction and a {\em 2D mask matching and refining} stage to produce multi-view consistent 2D masks based on the 3D masks for instance field supervision.



First, the {\em NeRF-RCNN}, which is extented from NeRF-RPN~\cite{nerf-rpn}, takes the extracted radiance and density field of the pre-trained NeRF and output 3D bounding boxes, class labels, and discrete 3D masks for each detected object in the NeRF. Then, given a set of multi-view but possibly \textit{inconsistent} 2D panoptic segmentation maps of the scene produced by existing methods, we project the 3D masks from the same camera poses to match the same instance across different views. The multi-view \textit{consistent} 2D segmentation can then be used to train the instance field component and produce a continuous 3D segmentation using instance field representation.


\subsection{Instance-NeRF}
\label{sec:Mask-NeR}
\begin{figure}[t]
\centering
    \includegraphics[width=1\linewidth]{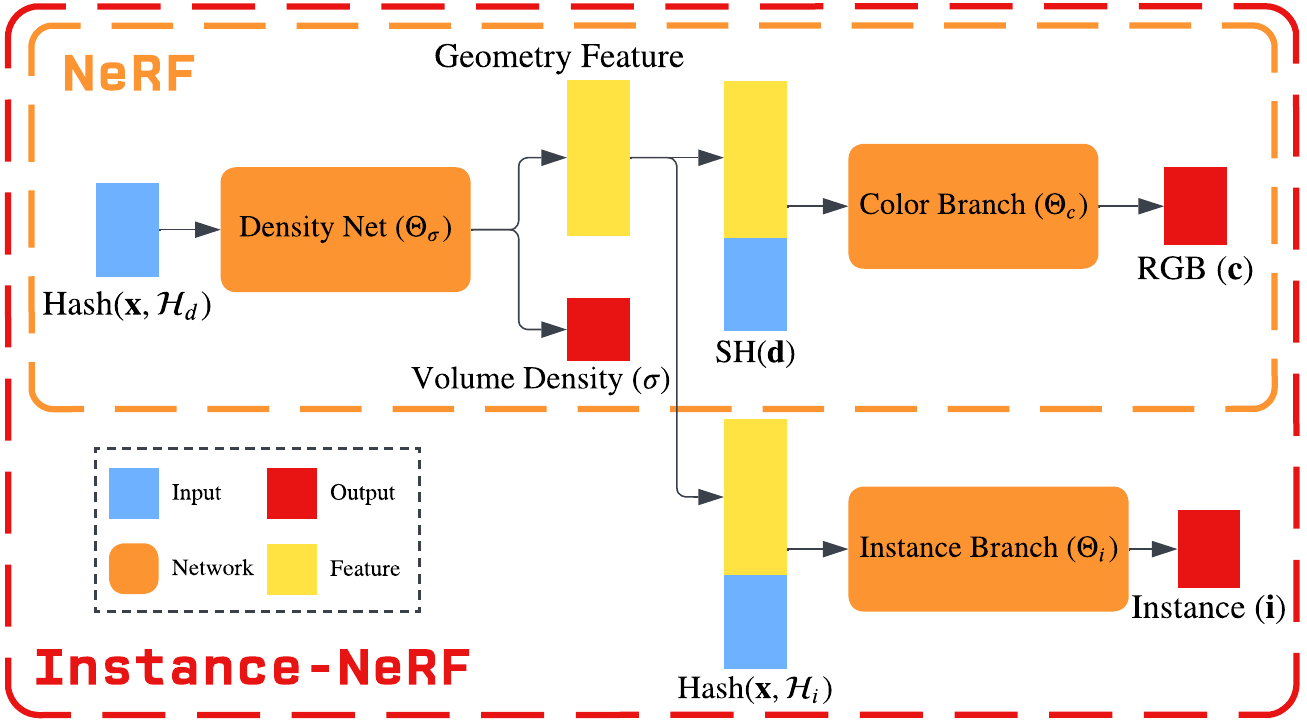}
    \vspace{-0.1in}
    
    \caption{\textbf{\inerf Architecture.} \inerf consists of density net $\Theta_\sigma$, color branch $\Theta_c$, and instance branch $\Theta_i$. Given the input spatial position $\mathbf{x}=(x,y,z)$ and a viewing direction $\mathbf{d}=(\phi, \theta)$, compared to the traditional \nerf, \inerf can predict a view-independent instance distribution $\mathbf{i}$ with its additional instance branch. In this paper, we adopt the Multi-resolution Hash Encoding (Hash) in~\cite{mueller2022instant} for position encoding and Spherical Harmonics (SH) in~\cite{yu2021plenoctrees} for viewing direction encoding. Note that the hash grid $\mathcal{H}_i$ used for mask net is different from the one $\mathcal{H}_\sigma$ used for density net.}\vspace{-0.15in}
    \label{fig:instance_nerf}
\end{figure}
\subsubsection{Model Architecture}
\inerf optimizes a neural radiance field that maps a 3D position $\mathbf{x}=(x,y,z)$ and a viewing direction $\mathbf{d}=(\phi, \theta)$ to a volume density $\sigma$, RGB radiance $\mathbf{c}=(r,g,b)$, and an instance label distribution over $L$ labels including the background label. 
As shown in Figure~\ref{fig:instance_nerf}, \inerf is parameterized by three components -- density net $\Theta_\sigma$, color branch $\Theta_c$, and instance branch $\Theta_i$. $\Theta_\sigma, \Theta_c$ constitute the original \nerf, and are assumed to be pre-trained with posed RGB images. The pre-trained \nerf can be formulated as
\begin{equation}
    \mathcal{F}_{\Theta_\sigma, \Theta_c}(\mathbf{x}, \mathbf{d}) = (\sigma, \mathbf{c}),
\end{equation}
and the Instance Field can be formulated as:
\begin{equation}
    \mathcal{G}_{\Theta_\sigma, \Theta_i}(\mathbf{x}) = \mathbf{i},
\end{equation}
where $\mathbf{i}$ is a $L$-dimensional vector with its first dimension representing the background.

We follow the hierarchical stratified sampling method in \cite{mildenhall2020nerf} to render the RGB color, depth, and instance label for a single pixel. Specifically, for each pixel, we formulate a ray $\mathbf{r}(t)=\mathbf{o}+t \mathbf{d}$ emitted from the center of the projection camera to that pixel, and select $K$ quadrature points $\left\{t_k\right\}_{k=1}^K$ on the ray $\mathbf{r}$ within the traceable volume of Instance-NeRF. The numerical quadrature to accumulate the expected color $\hat{\mathbf{C}}(\mathbf{r})$, expected depth $\hat{\mathbf{D}}(\mathbf{r})$ and expected instance logits $\hat{\mathbf{I}}(\mathbf{r})$ for each pixel are respectively given by:
\begin{equation}
    \hat{\mathbf{C}}(\mathbf{r})=\sum_{k=1}^K \hat{T}\left(t_k\right) \alpha\left(\sigma\left(t_k\right) \delta_k\right) \mathbf{c}\left(t_k\right),
\end{equation}
\begin{equation} \label{eq:logistic}
    \hat{\mathbf{D}}(\mathbf{r})=\sum_{k=1}^K \hat{T}\left(t_k\right) \alpha\left(\sigma\left(t_k\right) \delta_k\right) t_k ,
\end{equation}
\begin{equation} \label{eq:logistic}
    \hat{\mathbf{I}}(\mathbf{r})=\sum_{k=1}^K \hat{T}\left(t_k\right) \alpha\left(\sigma\left(t_k\right) \delta_k\right) \mathbf{i}\left(t_k\right),
\end{equation}
where
\begin{equation}
    \begin{array}{l}
        \hat{T}\left(t_k\right)=\exp \left(-\sum_{a=1}^{k-1} \sigma\left(t_a\right) \delta_a\right),  \\ 
        \alpha(x)=1-\exp (-x), \\
        \delta_k=t_{k+1}-t_k.
    \end{array}
\end{equation}
We discuss the NeRF pre-training and Instance Field training in sections~\ref{sec:nerf_pretraining} and \ref{sec:mask_nerf_training}, respectively. 

\subsubsection{Pre-training Density and Color}
\label{sec:nerf_pretraining}
The density and radiance components of \inerf can be implemented and trained similarly as a regular NeRF, and the implementation is largely orthogonal to the Instance Field component. Posed RGB images are used to train the density net $\Theta_\sigma$ and color branch $\Theta_c$ with the appearance loss $\mathcal{L}_p$:
\begin{equation}
    \mathcal{L}_p=\frac{1}{\mathcal{R}}\sum_{\mathbf{r} \in \mathcal{R}}\left\|\hat{\mathbf{C}}(\mathbf{r})-\mathbf{C}(\mathbf{r})\right\|_2^2,
\end{equation}
where $\mathcal{R}$ are the sampled rays within a training batch, $\mathbf{C}(\mathbf{r})$ and $\hat{\mathbf{C}}(\mathbf{r})$ are respectively the  ground truth and predicted RGB color for ray $\mathbf{r}$. In this work, we assume a well-trained $\Theta_c$ and $\Theta_\sigma$ are given and focuses mainly on the Instance Field training.


\subsubsection{Instance Field Training}
\label{sec:mask_nerf_training}
The instance field is trained with multi-view {\em consistent} 2D instance segmentation maps, during which $\Theta_\sigma$ and $\Theta_c$ will be fixed and only $\Theta_i$ will be optimized. Given a set of multi-view consistent 2D instance segmentation maps, we train the instance branch with multi-class cross-entropy loss $\mathcal{L}_i$:

\begin{equation}
    \mathcal{L}_i=-\frac{1}{\mathcal{R}L}\sum_{\mathbf{r} \in \mathcal{R}}\sum_{l=1}^L p^l(\mathbf{r}) \log \hat{p}^l(\mathbf{r}),
\end{equation}
where $p^l(\mathbf{r})$ and $\hat{p}^l(\mathbf{r})$ are respectively the ground truth multi-class instance probability for class $l$ and volume predictions for ray $\mathbf{r}$ given by applying softmax to Eq.~\ref{eq:logistic}.  

As the input 2D segmentation maps are generated by existing 2D segmentation methods, it cannot produce 3D geometry-consistent masks for different views. To compensate this issue and to encourage the smoothness of the instance label prediction, we take advantage from the prior that instance maps are generally smooth over local regions, and add an instance regularization loss  $\mathcal{L}_r$ adopted from~\cite{Niemeyer2021Regnerf}:

\begin{align}
\begin{split}
 \mathcal{L}_r = \frac{1}{\mathcal{R}L}\sum_{\mathbf{r}_{i,j} \in \mathcal{R}}\sum_{l=1}^L  \left(\hat{\mathbf{I}}(\mathbf{r}_{i,j}) - \hat{\mathbf{I}} (\mathbf{r}_{i,j+1}) \right)^2 w_{i,j}  + \\ \left(\hat{\mathbf{I}}(\mathbf{r}_{i,j}) - \hat{\mathbf{I}}(\mathbf{r}_{i+1,j}) \right)^2 h_{i,j},
\end{split}
\end{align}
\begin{equation}
    w_{i,j} = \frac{\delta_w (i,j)}{\sum_{\mathbf{r}_{i',j'} \in \mathcal{R} } \delta_w (i',j') },
\end{equation}
\begin{equation}
    \delta_w (i,j) = \exp(-(\hat{\mathbf{D}}(\mathbf{r}_{i,j}) - \hat{\mathbf{D}} (\mathbf{r}_{i,j+1}) )^2),
\end{equation}
where $w_{i,j}$ is the weight determined by the similarity of NeRF depth $\hat{\mathbf{D}}$ between two sample rays, and likewise for $h_{i,j}$. $\mathbf{r}_{i,j}$ is the ray passing through pixel $(i,j)$ of the image. Recall that depth information is obtained from the pre-trained density branch. Similar depths of adjacent pixels imply that the pixels might be on the same surface in 3D space, and likely belong to the same object. 
Thus, this regularization loss encourages the model to predict the same instance label for them and generate smoother segmentation for each object. 

The total loss $\mathcal{L}$ for training the instance field is:
\begin{equation}
\label{eq:total_loss}
    \mathcal{L} = \mathcal{L}_p + \lambda \mathcal{L}_i,
\end{equation}
where $\lambda$ is a hyperparameter.

\subsection{NeRF-RCNN}
\label{sec:NeRF-RCNN}
{\em NeRF-RCNN} takes as input a pre-trained NeRF and outputs 3D Axis-Aligned Bounding Boxes (AABB), class labels, and discrete 3D masks of the detected objects. This network extends NeRF-RPN~\cite{nerf-rpn} by appending two additional detection heads: a box prediction head (MLP) and a mask prediction head (3D CNN), much similar in spirit as \cite{he2017mask}. As shown in Figure~\ref{fig:main}, we first uniformly sample the appearance and density on a grid that covers the full traceable volume of the pre-trained \inerf model, following the sampling method in \cite{nerf-rpn}. The NeRF-RPN takes the sampled grid as input, and outputs proposals and the feature pyramid. We use 3D RoIAlign to obtain the feature within the RoI of each output proposal. The box prediction head takes RoI features as input and outputs class logits and AABB regression offset, while the mask prediction head takes RoI features as input and outputs 3D masks for each RoI.

The {\em NeRF-RCNN} can be trained in an end-to-end manner with sampled appearance and density grids as input, along with ground truth bounding boxes and 3D instance segmentation masks, by reducing the multi-class cross-entropy loss for class prediction, the smooth L1 loss for bounding box regression, and the binary cross-entropy loss for 3D mask prediction. In practice, we adopt the pre-trained NeRF-RPN, and optimize the classification head and mask head side by side. In our experiments, we train the {\em NeRF-RCNN} with a large NeRF segmentation dataset built on 3D-FRONT.

\subsection{2D Mask Matching and Refinement}
\label{sec:refinement}
It is infeasible to adopt a 2D instance segmentation model to perform segmentation on multi-view images and directly use the results to supervise instance field training. The main reason lies in the fact that 2D instance segmentation models generally do not guarantee any correspondence between the appearance of the same instance in different views,  including the assigned instance IDs, predicted class labels, and the masks. The same object is likely to be assigned with different instance IDs in different images, making it difficult for direct use in instance field training. Multi-view consistency within 2D instance masks are thus crucial for the quality of the instance field.

To address this consistency issue, we utilize the 3D coarse mask produced by NeRF-RCNN to align masks in different images that correspond to the same object. We further propose an iterative refinement approach by feeding the intermediate results of \inerf to an existing mask refinement model, and use the refined masks to train the final instance field.


\subsubsection{2D Mask Matching}
\label{sec:mask_predict}
To obtain the initial 2D segmentation, we use a Mask2Former~\cite{cheng2021mask2former} panoptic segmentation model pretrained on the COCO~\cite{mscoco} dataset. We create a class mapping from COCO to the dataset we use, based on which we filter out masks that belong to background. To deal with prediction inconsistency across images, we match each Mask2Former predicted 2D mask with a 3D instance detected by NeRF-RCNN. To achieve this, we first project the 3D masks from NeRF-RCNN to the corresponding image spaces, and compute the Intersection over Union (IoU) between each pair of predicted 2D mask and projected 2D mask. Each predicted mask is then assigned to the instance with highest mask IoU. Those predicted masks that do not have an IoU greater than 0.05 with any projected masks are treated as unlabeled area, which do not participate in instance NeRF training. After this process, we have consistently annotated 2D instance masks from multi-view images, which will be used to optimize the instance field.  

\subsubsection{2D Mask Refinement}
Although 2D-to-3D matching emancipates the model from inconsistency, uniformly sampled images, which frequently contain objects under partial occlusion
can easily mislead the prediction model pre-trained on general datasets, resulting in missing or erroneous mask prediction in different views. 
An instance field trained directly on the 2D masks therefore suffers from conflicting 2D masks of different views, and usually produces fragmented segmentation results. To alleviate this issue, we adopt a pre-trained 2D mask refinement model, CascadePSP~\cite{cheng2020cascadepsp}, to refine the preliminary instance segmentation results of \inerf. Instead of directly feeding Mask2Former output into the refinement network, we use \inerf rendering results, because \inerf can enhance single-view results by fusing multi-view information, filling part of the area of incomplete or missing mask prediction in certain views. We split the instances in each preliminary \inerf mask and refine them independently, the results of which are superimposed to supervise the final instance field training.



\newcommand{\snerf}{Semantic-NeRF}
\newcommand{\dmnerf}{DM-NeRF}
\newcommand\cwidth{0.25\linewidth}
\newcommand\resultWidth{0.03cm}
\newcommand\resultHeight{-0.2cm}
\begin{figure*}[h]
    \centering
    \captionsetup[subfloat]{position=top}




    \subfloat[Mask2Former]{\includegraphics[width = 0.18\linewidth]{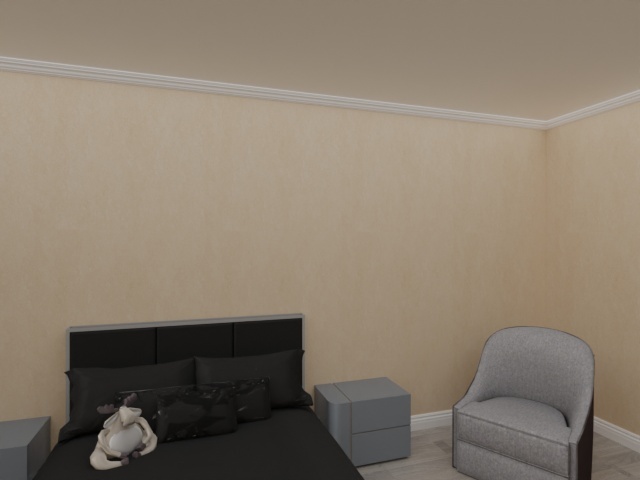}}\hspace{\resultWidth}
    \subfloat[Semantic-NeRF]{\includegraphics[width = 0.18\linewidth]{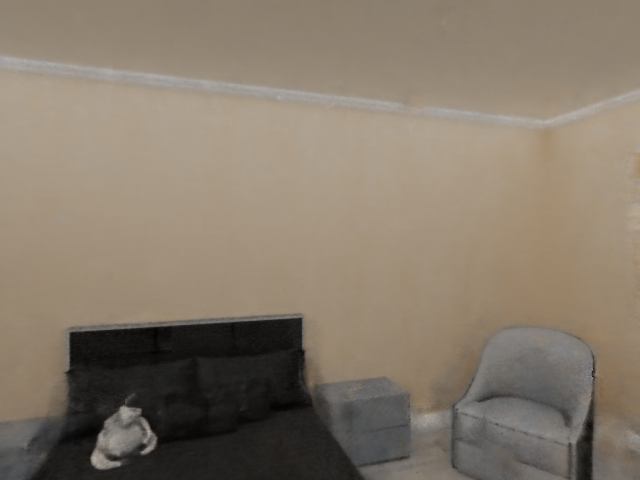}}\hspace{\resultWidth}
    \subfloat[DM-NeRF]{\includegraphics[width = 0.18\linewidth]{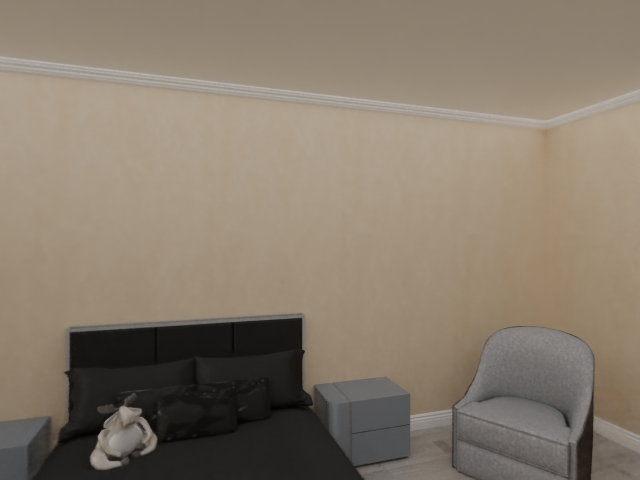}}\hspace{\resultWidth}
    \subfloat[Ours]{\includegraphics[width = 0.18\linewidth]{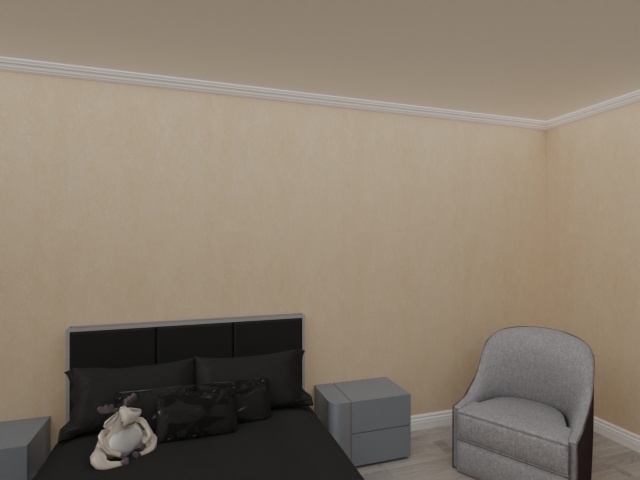}}\hspace{\resultWidth}    
    \subfloat[GT]{\includegraphics[width = 0.18\linewidth]{figs/3dfront_0091_00/3dfront_0091_00_0168.jpg}}

    \vspace{\resultHeight}

    \subfloat{\includegraphics[width = 0.18\linewidth]{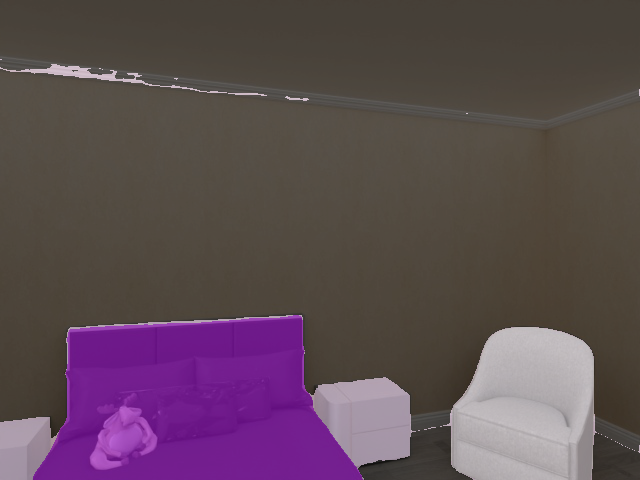}}\hspace{\resultWidth}
    \subfloat{\includegraphics[width = 0.18\linewidth]{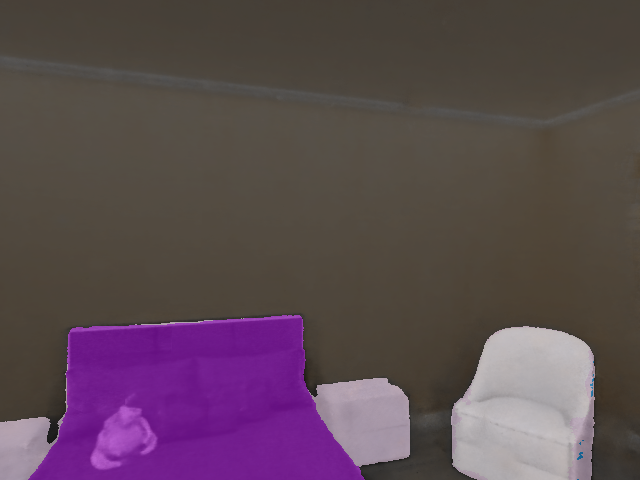}} \hspace{\resultWidth}
    \subfloat{\includegraphics[width = 0.18\linewidth]{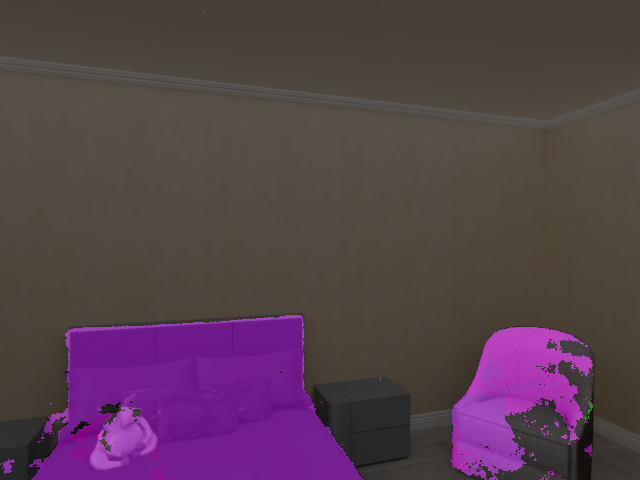}}\hspace{\resultWidth}
    \subfloat{\includegraphics[width = 0.18\linewidth]{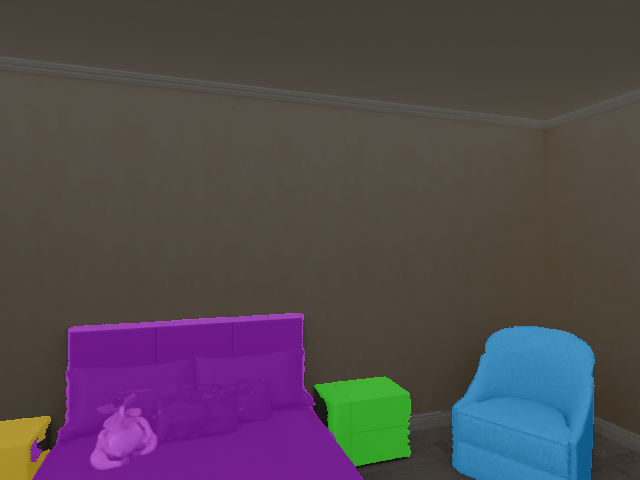}}\hspace{\resultWidth}    
    \subfloat{\includegraphics[width = 0.18\linewidth]{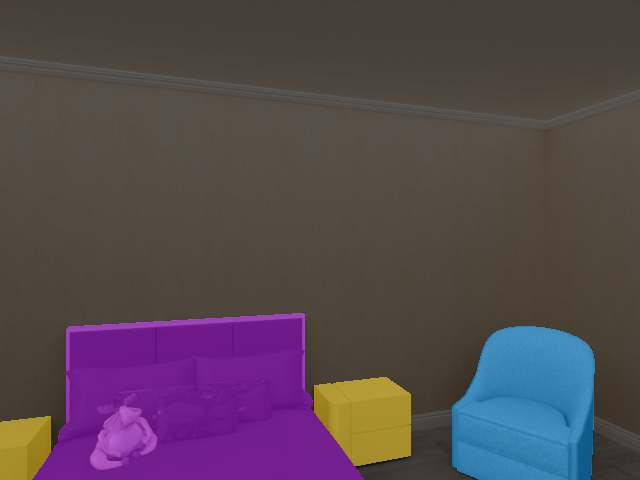}}

    \vspace{\resultHeight}

	\subfloat{\includegraphics[width = 0.18\linewidth]{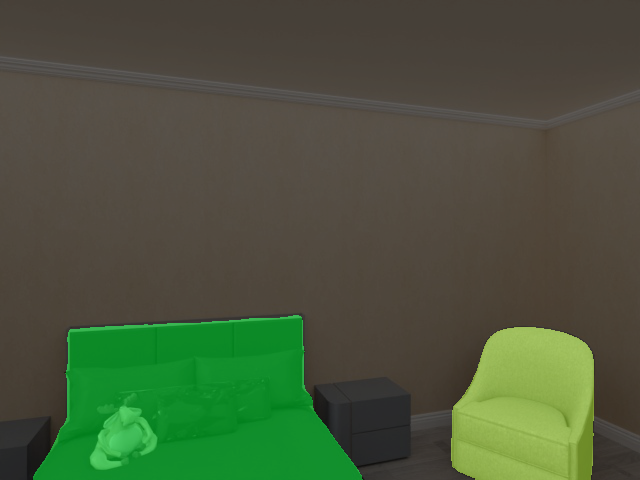}}\hspace{\resultWidth}
    \subfloat{\includegraphics[width = 0.18\linewidth]{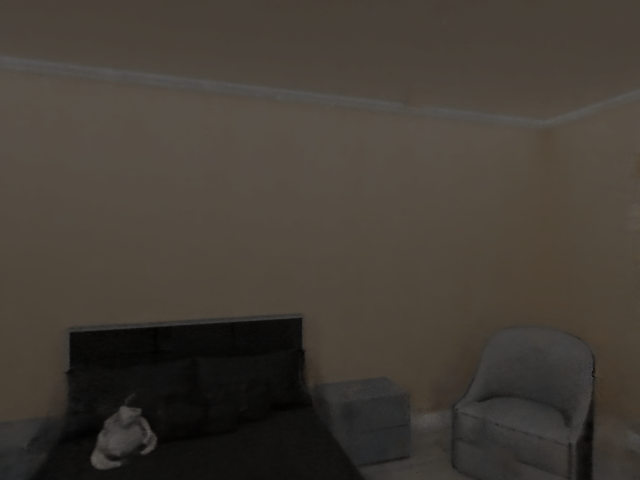}} \hspace{\resultWidth}
    \subfloat{\includegraphics[width = 0.18\linewidth]{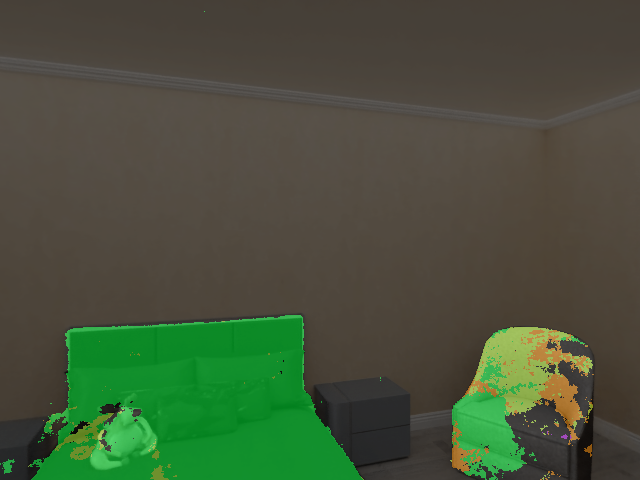}}\hspace{\resultWidth}
	\subfloat{\includegraphics[width = 0.18\linewidth]{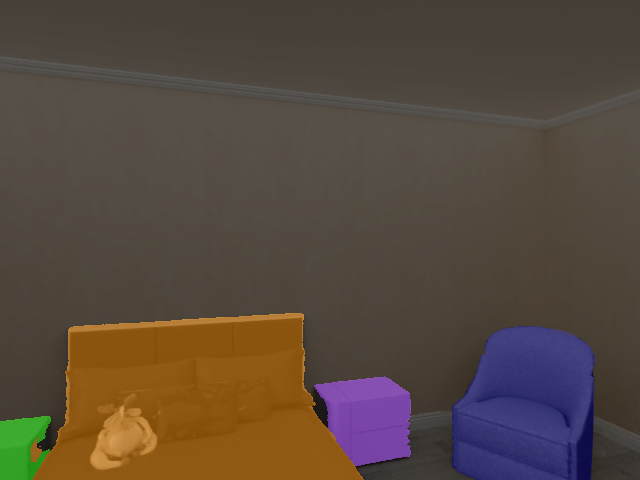}}\hspace{\resultWidth}    
    \subfloat{\includegraphics[width = 0.18\linewidth]{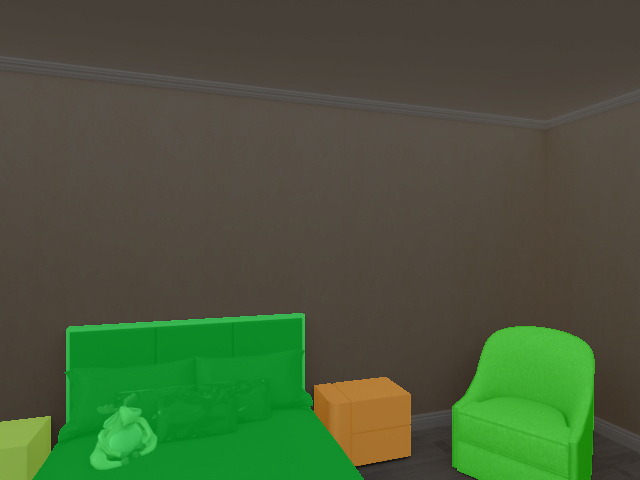}}

\caption{\textbf{Comparison.} This figure illustrates the comparison with other state-of-art methods. Rows from top to bottom are i. ground truth RGB images or the rendered RGB image by the models, ii. semantic segmentation, and iii. instance segmentation. The instance segmentation results from Semantic-NeRF are left empty as it does not produce instance-level information.}

    \label{fig:comparison}
\end{figure*}

\renewcommand\resultWidth{0.03cm}
\renewcommand\resultHeight{-0.2cm}
\begin{figure*}[ht]
    \centering
    \captionsetup[subfloat]{position=top}

    \subfloat[GT Semantic]{\includegraphics[width = 0.18\linewidth]{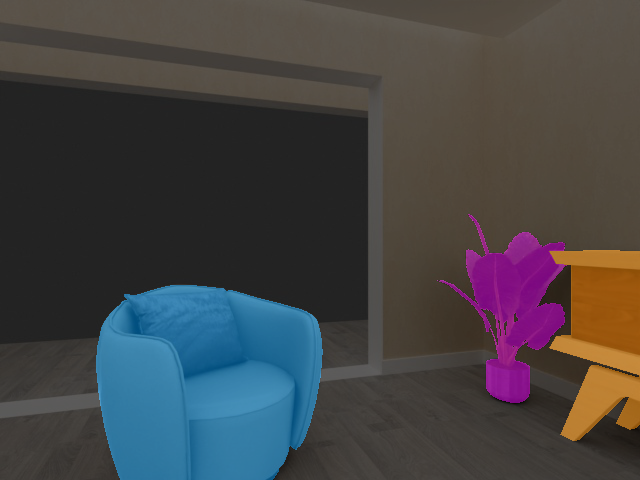}}\hspace{\resultWidth}
    \subfloat[Predicted Semantic]{\includegraphics[width = 0.18\linewidth]{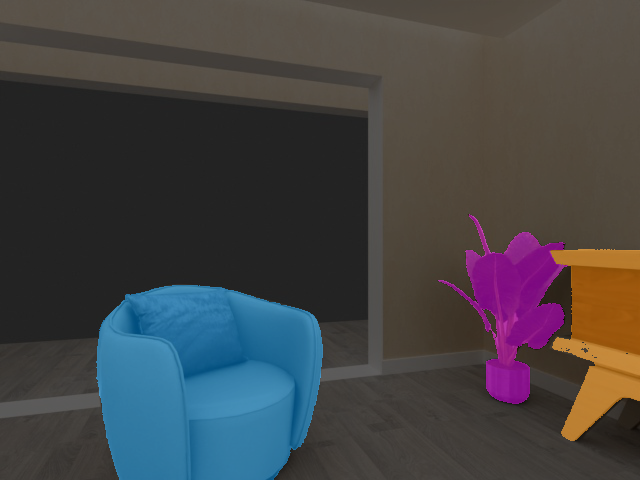}}\hspace{\resultWidth}
    \subfloat[GT Instance]{\includegraphics[width = 0.18\linewidth]{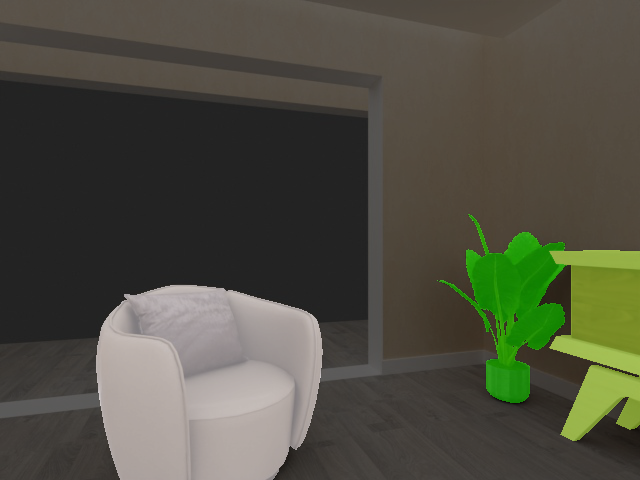}}\hspace{\resultWidth}
    \subfloat[Predicted Instance]{\includegraphics[width = 0.18\linewidth]{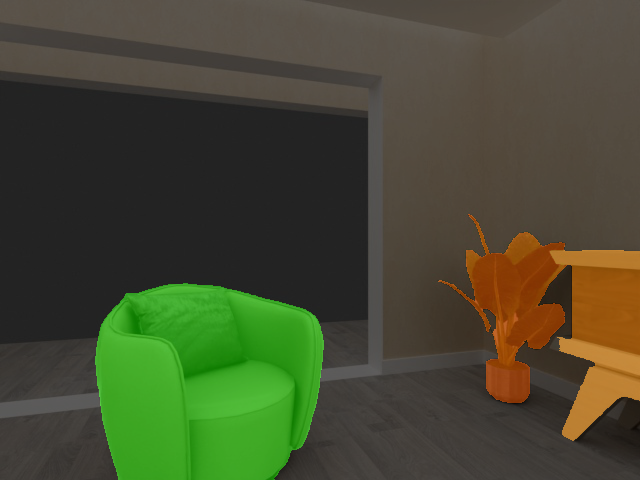}}\hspace{\resultWidth}    
    \subfloat[GT RGB]{\includegraphics[width = 0.18\linewidth]{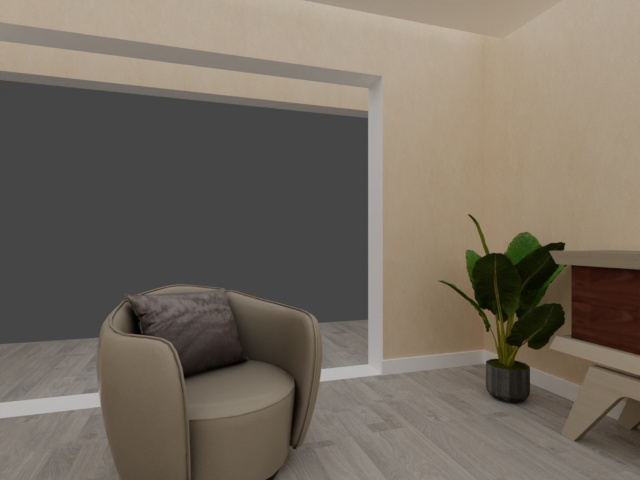}}

    \vspace{\resultHeight}

	\subfloat{\includegraphics[width = 0.18\linewidth]{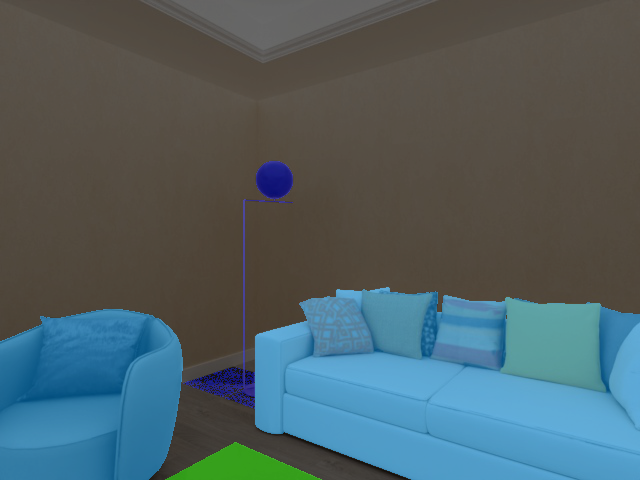}}\hspace{\resultWidth}
    \subfloat{\includegraphics[width = 0.18\linewidth]{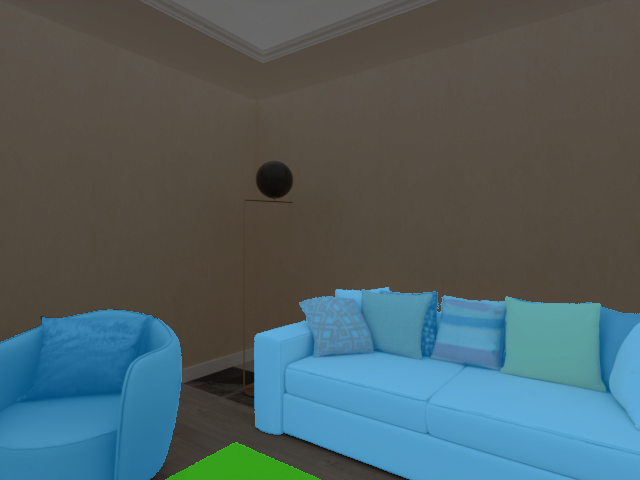}} \hspace{\resultWidth}
    \subfloat{\includegraphics[width = 0.18\linewidth]{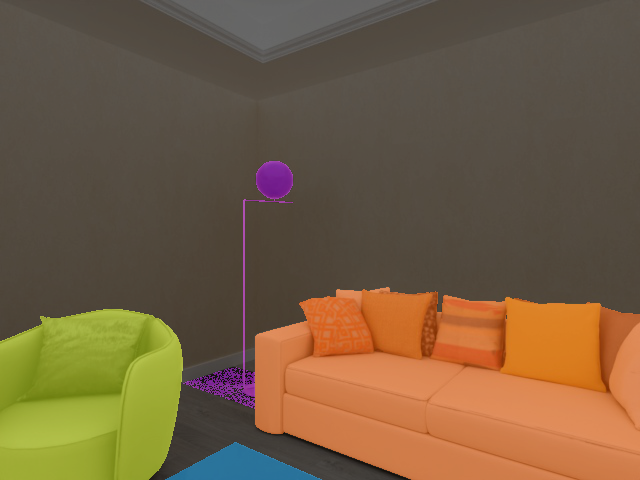}}\hspace{\resultWidth}
	\subfloat{\includegraphics[width = 0.18\linewidth]{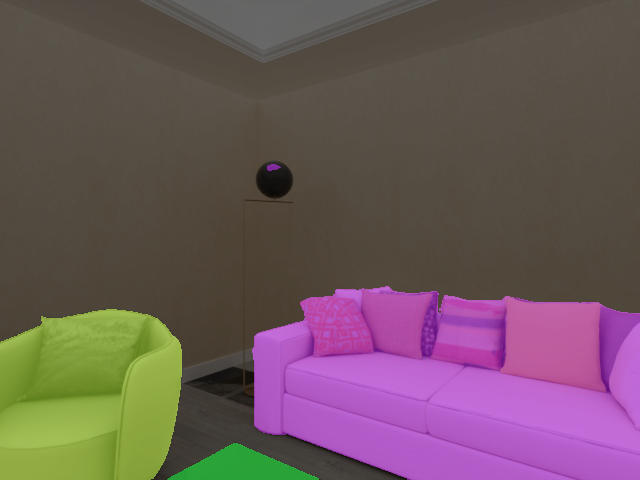}}\hspace{\resultWidth}    \subfloat{\includegraphics[width = 0.18\linewidth]{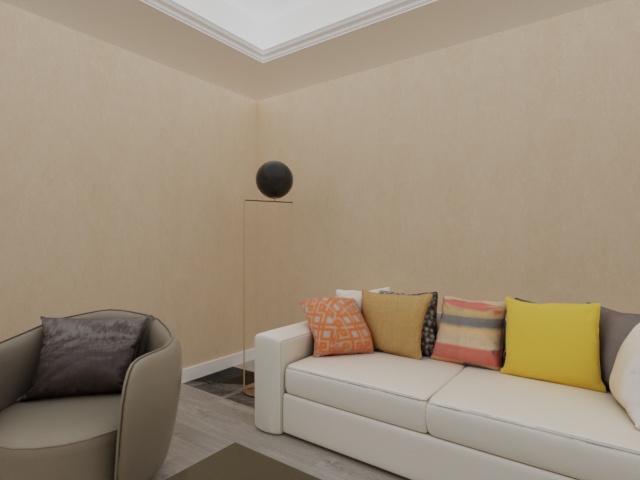}}

    \vspace{\resultHeight}

	\subfloat{\includegraphics[width = 0.18\linewidth]{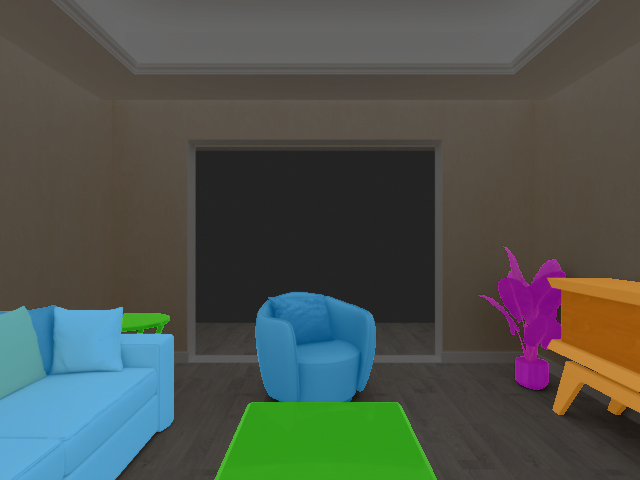}}\hspace{\resultWidth}
    \subfloat{\includegraphics[width = 0.18\linewidth]{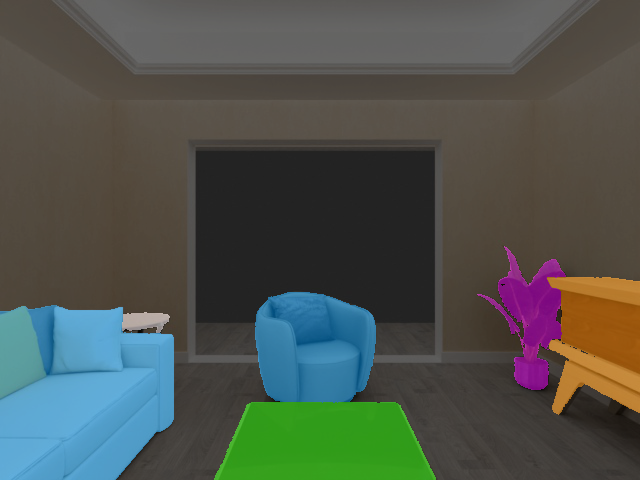}} \hspace{\resultWidth}
    \subfloat{\includegraphics[width = 0.18\linewidth]{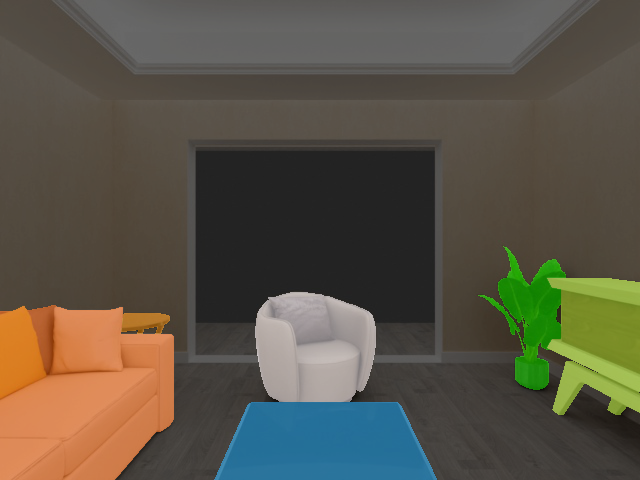}}\hspace{\resultWidth}
	\subfloat{\includegraphics[width = 0.18\linewidth]{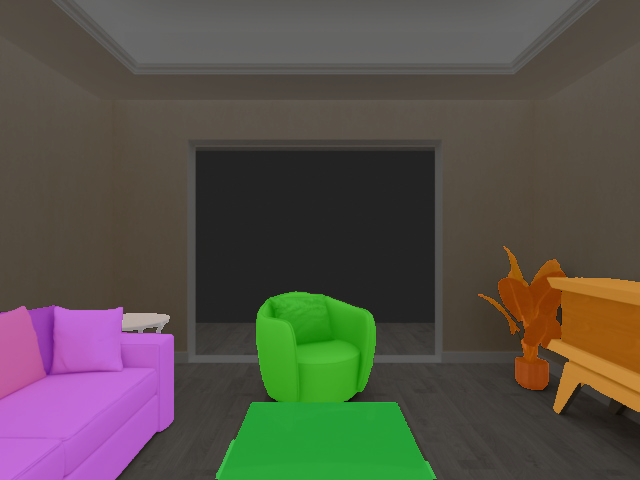}}\hspace{\resultWidth}    \subfloat{\includegraphics[width = 0.18\linewidth]{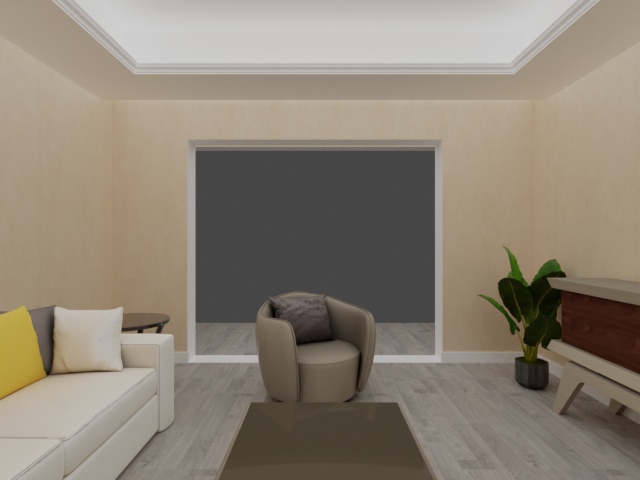}}
	\vspace{-0.1cm}
    \caption{\textbf{Results.} This figure shows our multi-view results from a single scene. Our method achieves high consistency across images (Column (b) and Column (d)). }
    \vspace{0.2cm}
    \label{fig:results}
\end{figure*}

\section{Experiments}

\subsection{Training and Testing}
\noindent\textbf{Training.} 
In our model, only {\em NeRF-RCNN} requires a large number of scenes for training for generalization, while the instance field training are scene-specific and the models used in 2D mask matching and refinement are pre-trained. Following NeRF-RPN~\cite{nerf-rpn} strategy and configuration, NeRF-RCNN is trained on 3D-FRONT~\cite{fu20213d} NeRF segmentation dataset. We follow the dataset creation approach in~\cite{nerf-rpn} and extend the dataset to include 1016 scenes, 814 of which are used for training {\em NeRF-RCNN} and 101 scenes are used for validation and testing, respectively. We choose VGG19, which performs best on object detection, as our backbone. The weights of the three losses mentioned in Section \ref{sec:NeRF-RCNN} are all equal to 1.0. We apply random flipping and rotation by $\pi/2$ with probability 0.5 to the input as augmentation.

For the 2D panoptic segmentation model, we use the Mask2Former model pre-trained on COCO dataset, with Swin-L as the backbone. For the mask refinement, we adopt the publicly available CascadePSP model. No fine-tuning is performed for both methods on our dataset.
\vspace{2mm}
 
\noindent\textbf{Testing}
We test our model on 8 scenes from 3D-FRONT in total. We use a non-maximum suppression (NMS) threshold of 0.3 for NeRF-RPN and 0.15 for NeRF-RCNN. We use a score threshold of 0.5 to filter the NeRF-RCNN detection. The threshold for Mask2Former segmentation is also 0.5. For the second training stage of \inerf, we implement our model based on~\cite{torch-ngp}.
The hyperparameter $\lambda$ in Eq.~(\ref{eq:total_loss}) is 1.0. For each scene, we train the radiance and density field on a single NVIDIA 1080ti for 30k iterations. The instance field is then trained for 25k steps to obtain the intermediate segmentation results for refinement. We train the instance field for another 20k steps using the refined 2D masks. Figure~\ref{fig:results} illustrates the high-quality 3D instance segmentation results. 


\begin{table}[]
\centering
\resizebox{0.6\linewidth}{!}{
\begin{tabular}{lll}
\hline
Methods & mIoU $\uparrow$ & PQ $\uparrow$ \\ \hline
Mask2Former & 31.6 & 22.5 \\
Semantic-NeRF & 35.0 & - \\
DM-NeRF & 11.3 & 5.4 \\
Ours & \textbf{43.0} & \textbf{32.4} \\ \hline
\end{tabular}
}
\vspace{-0.05in}
\caption{Comparison results of mean Intersection over
Union (mIoU) and Panoptic Quality (PQ) on 3D-FRONT dataset.}

\label{tab:comparison}
\end{table}


\begin{table}[]
\centering
\resizebox{0.65\linewidth}{!}{
\begin{tabular}{lcll}
\cline{1-4}
$\mathcal{L}_i$ & Mask Refinement & mIoU & PQ \\ \cline{1-4} 
 &  & 39.8 & 27.6 \\
\checkmark &  & 40.8 & 29.6 \\
\checkmark & \checkmark & \textbf{43.0} & \textbf{32.4} \\ \cline{1-4} 
\end{tabular}
}
\vspace{-0.05in}
\caption{Ablation over instance regularization loss and 2D mask refinement on 3D-FRONT dataset.}
\label{tab:ablation}
\end{table}

\subsection{Metrics}
We evaluate the semantic segmentation performance of \inerf on novel views using mean Intersection over Union (mIoU) and instance segmentation performance using Panoptic Quality (PQ). The semantic segmentation is acquired by mapping the classes of the detected instances. We use PQ for evaluating instance segmentation as not all compared methods produce instance confidence scores, hence we cannot use common metrics such as mean Average Precision. Classes corresponding to background are filtered for PQ calculation. Note that we compute PQ in a conventional approach, where the consistency of instances across multiple frames is not taken into account. This allows us to compare with Mask2Former even it does not produce multi-view consistent segmentation. However, PQ also fails to reflect our method's performance in producing consistent segmentation for different views.

\subsection{Comparison}
To our best knowledge, we are one of the first to propose 3D instance segmentation for \nerf without utilizing ground-truth information at test time, which can be generalized to different kinds of scenes and requires pre-trained \nerf only during inference. Thus, it is hard to find comparable methods with the same configuration. However, we still provide the comparison with some related methods mentioned in Section~\ref{sec:inst_seg} 

Semantic-NeRF~\cite{semantic_nerf} also uses a 2D pre-trained model but the method cannot be straight-forwardly modified for instance segmentation. DM-NeRF~\cite{wang2022dmnerf} can achieve consistency in 3D with the supervision of 2D ground truth instance segmentation. In our experiments, we compare our method with Semantic-NeRF for semantic segmentation, and with DM-NeRF for instance segmentation. For fairness, we implement them on torch-ngp~\cite{torch-ngp} and use Mask2former~\cite{cheng2021mask2former} to predict 2D masks for input images, supervising the \nerf training of these two methods. As Mask2Former do not guarantee multi-view instance consistency, we use the majority class for each instance to create the mapping to class semantics. Figure~\ref{fig:comparison} illustrates the comparison of the qualitative results and Table~\ref{tab:comparison} tabulates the quantitative comparison. The noisy instance masks from DM-NeRF verify the idea that multi-view consistent 2D masks are vital for high-quality instance field supervision, and the results of our method demonstrate the effectiveness of the mask matching process. 

\begin{table}[]

\begin{center}
    

\resizebox{0.70\linewidth}{!}{
\begin{tabular}{lcc}
    
\hline
             & Accuracy $\uparrow$   & mIoU $\uparrow$  \\ \hline
uORF~\cite{yu2022unsupervised}              & 75.68            &  37.85      \\
NeRF-SOS~\cite{fan2022nerf}                 & 75.56      & 57.61  \\
RFP~\cite{liu2022unsupervised}           & 66.63      & 47.66  \\
\emph{Ours}   & \textbf{95.48}  & \textbf{89.34}  \\ \hline
\end{tabular}
}
\end{center}

\vspace{-0.05in}
\caption{\textbf{Comparison with Unsupervised Methods}. To make fair comparison, the accuracy and mIoU are computed on binary classes over novel views (foreground/background). }
\vspace{-0.1in}
\label{tab:rebuttal-comparison}
\end{table}

\vspace{-0.15in}
\begin{figure}[ht]
    \centering
    \captionsetup[subfloat]{labelformat=empty}

    \subfloat{\includegraphics[width = 0.155\linewidth]{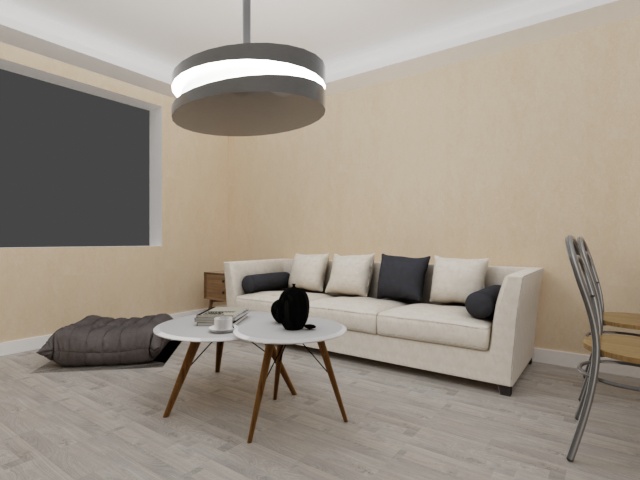}}\hspace{\width}
    \subfloat{\includegraphics[width = 0.155\linewidth]{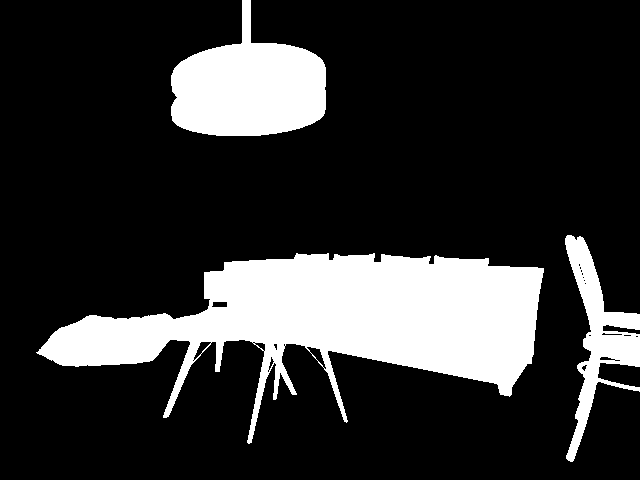}}\hspace{\width}
    \subfloat{\includegraphics[width = 0.155\linewidth]{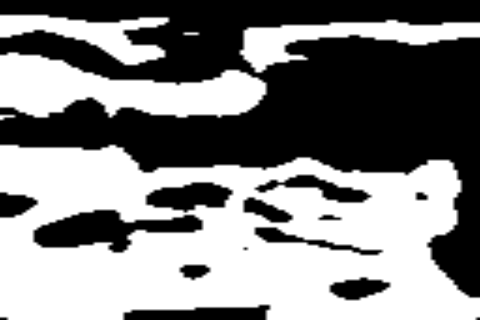}}\hspace{\width}
    \subfloat{\includegraphics[width = 0.155\linewidth]{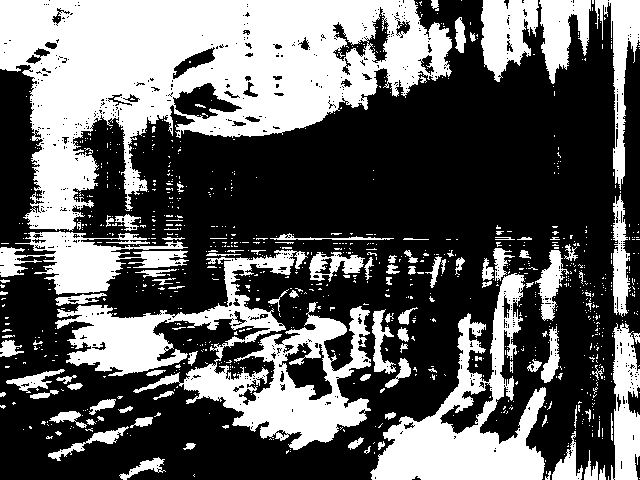}}\hspace{\width}
    \subfloat{\includegraphics[width = 0.155\linewidth]{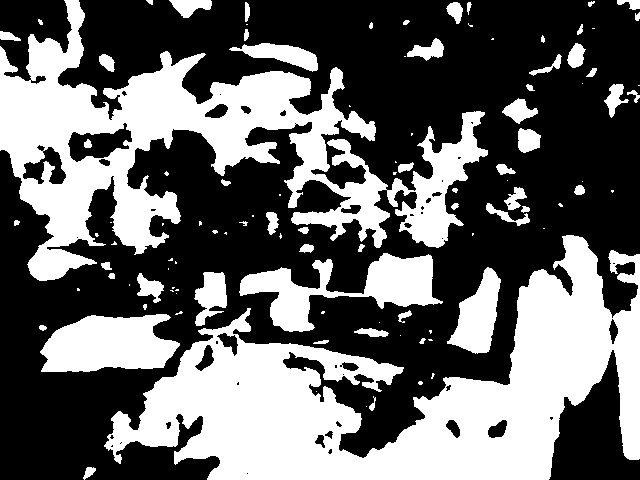}}\hspace{\width}
    \subfloat{\includegraphics[width = 0.155\linewidth]{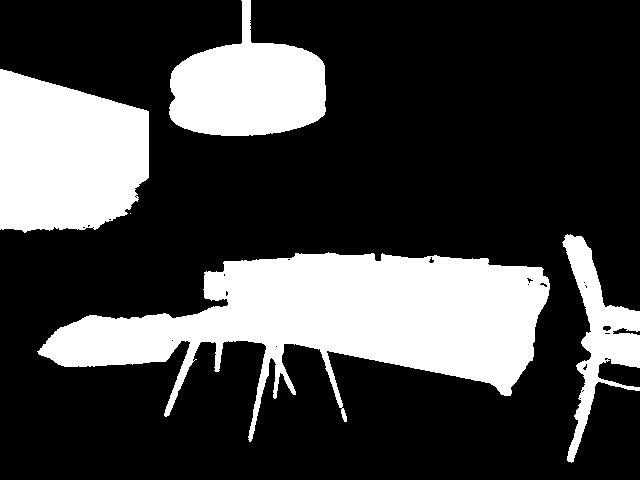}}
	
    \vspace{\height}

    \subfloat{\includegraphics[width = 0.155\linewidth]{figs/3dfront_0089_00/3dfront_0089_00_0258.jpg}}\hspace{\width}
    \subfloat{\includegraphics[width = 0.155\linewidth]{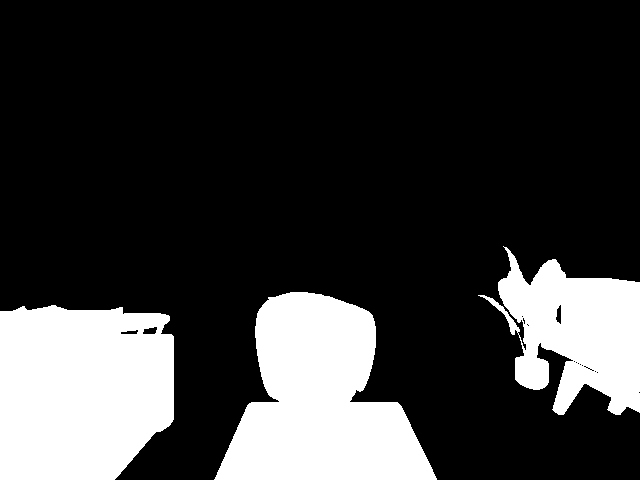}}\hspace{\width}
    \subfloat{\includegraphics[width = 0.155\linewidth]{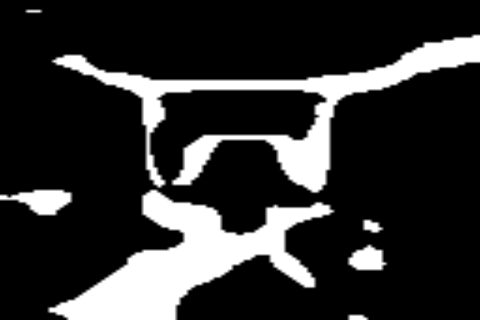}}\hspace{\width}
    \subfloat{\includegraphics[width = 0.155\linewidth]{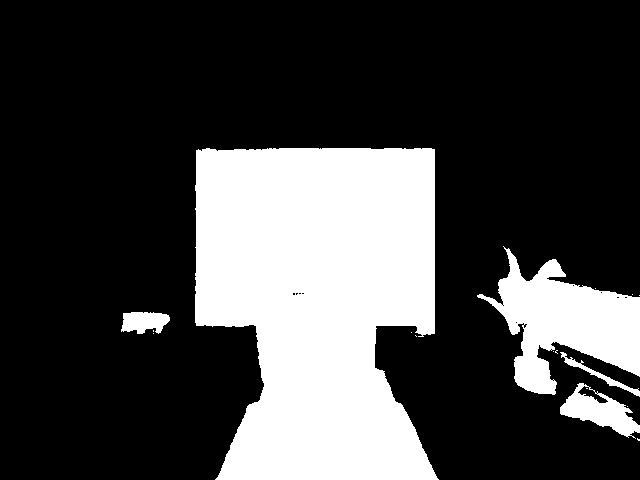}}\hspace{\width}
    \subfloat{\includegraphics[width = 0.155\linewidth]{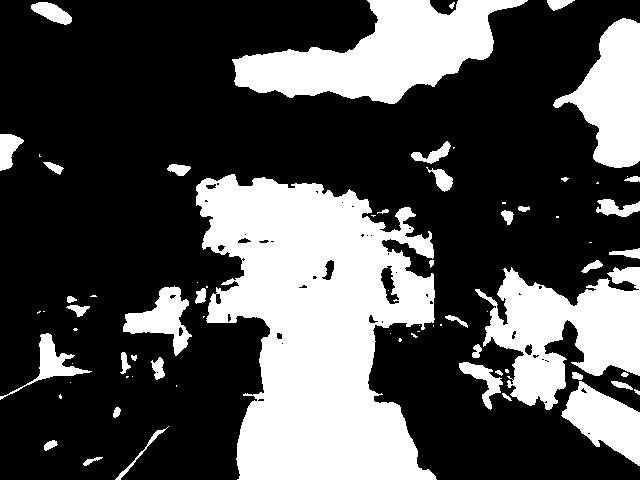}}\hspace{\width}
    \subfloat{\includegraphics[width = 0.155\linewidth]{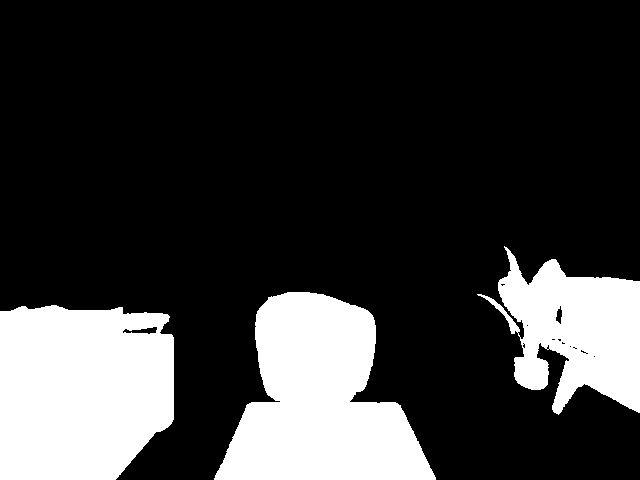}}
	
    \vspace{\height}

    \subfloat{\includegraphics[width = 0.155\linewidth]{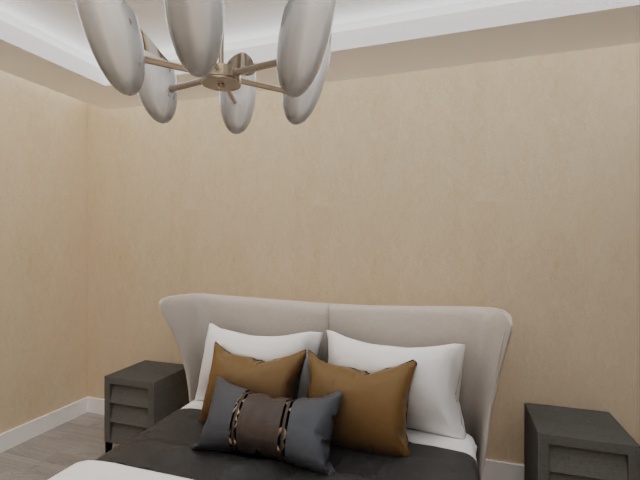}}\hspace{\width}
    \subfloat{\includegraphics[width = 0.155\linewidth]{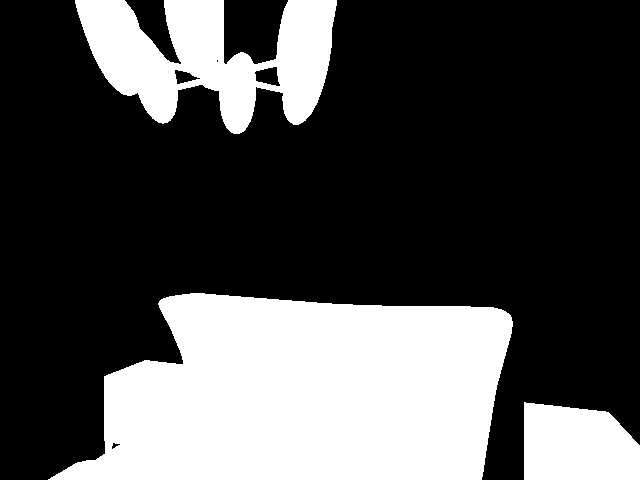}}\hspace{\width}
    \subfloat{\includegraphics[width = 0.155\linewidth]{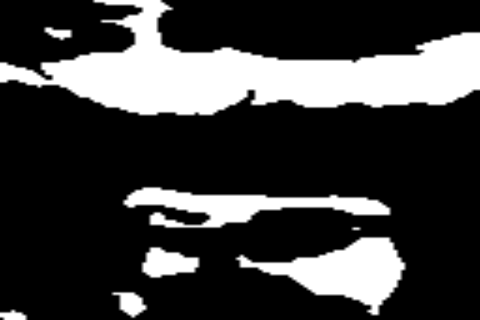}}\hspace{\width}
    \subfloat{\includegraphics[width = 0.155\linewidth]{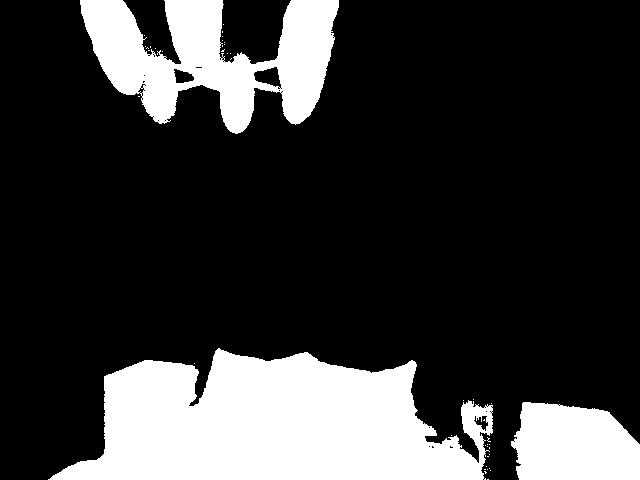}}\hspace{\width}
    \subfloat{\includegraphics[width = 0.155\linewidth]{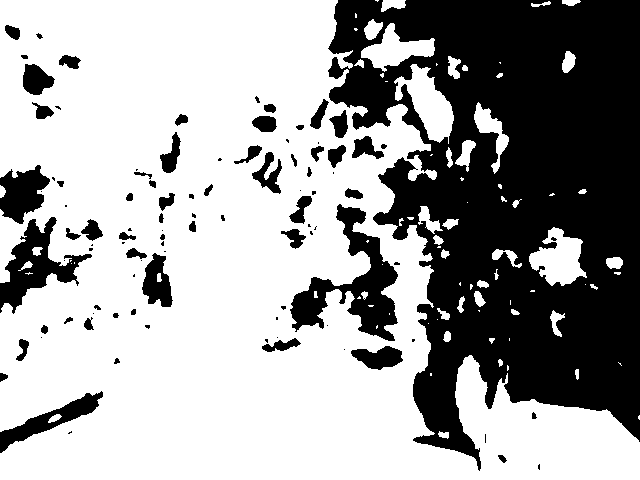}}\hspace{\width}
    \subfloat{\includegraphics[width = 0.155\linewidth]{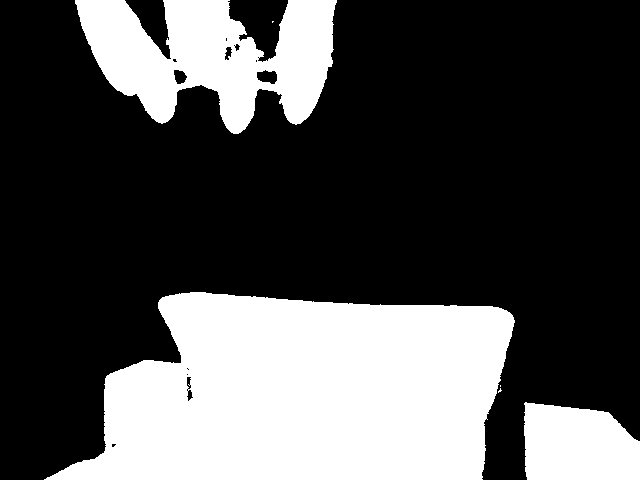}}
	
    \vspace{\height}

    \subfloat[RGB]{\includegraphics[width = 0.155\linewidth]{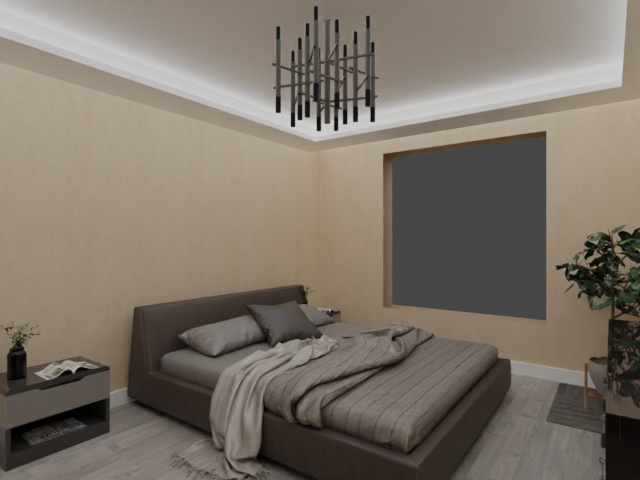}}\hspace{\width}
    \subfloat[GT Mask]{\includegraphics[width = 0.155\linewidth]{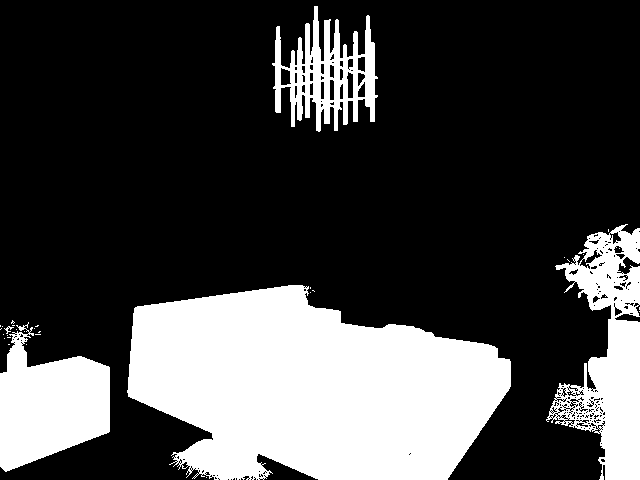}}\hspace{\width}
    \subfloat[uORF]{\includegraphics[width = 0.155\linewidth]{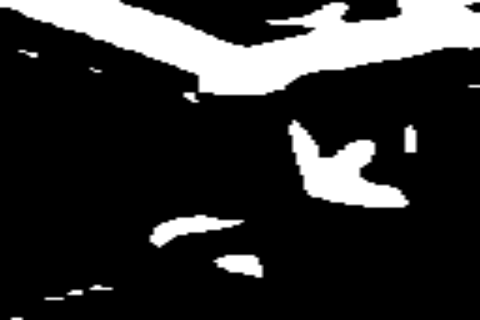}}\hspace{\width}
    \subfloat[NeRF-SOS]{\includegraphics[width = 0.155\linewidth]{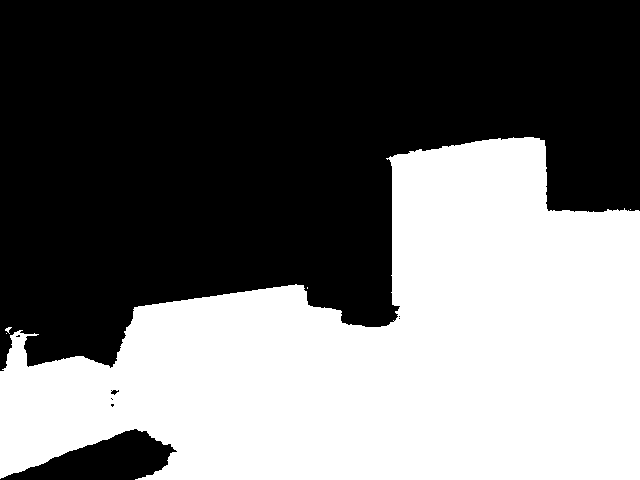}}\hspace{\width}
    \subfloat[RFP]{\includegraphics[width = 0.155\linewidth]{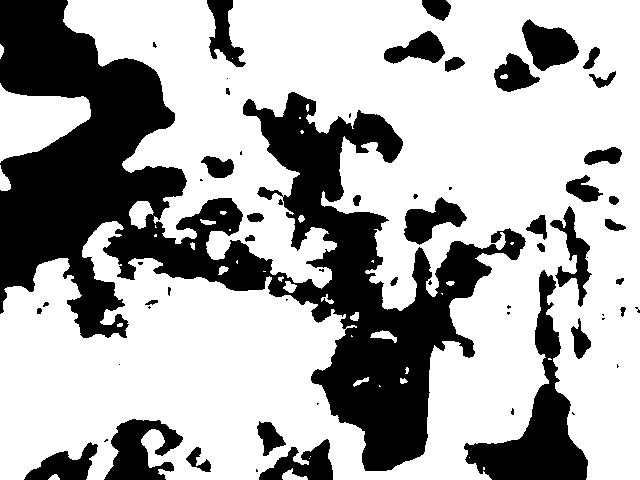}}\hspace{\width}
    \subfloat[\emph{Ours}]{\includegraphics[width = 0.155\linewidth]{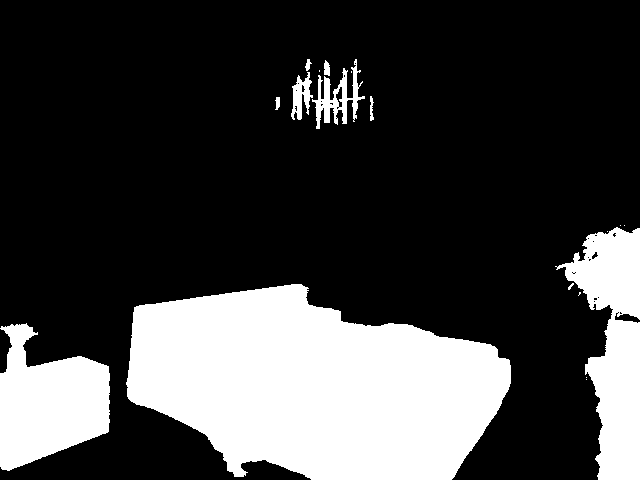}}
	
    \vspace{0.05in}
    \caption{\textbf{Qualitative Results of Comparison with Unsupervised Methods}. }
    \vspace{-0.05in}
    \label{fig:rebuttal-qualitative}
\end{figure}
\vspace{-2mm}

For unsupervised methods mentioned in section~\ref{sec:inst_seg}, none of them produces object classes. uORF~\cite{yu2022unsupervised} and ObSuRF~\cite{stelzner2021decomposing} are completely unsupervised and work on object-centric scenes, which are far different from our testing scenes. NeRF-SOS~\cite{fan2022nerf} and RFP~\cite{liu2022unsupervised} require a rough estimate of the number of objects in the scene, or multi-view consistent masks for multiple objects as an initial guess, and multi-object segmentation is not adequately addressed in their papers. Comparing these methods with ours is largely unfair. For reference, however, we provide the comparison results in Figure~\ref{fig:rebuttal-qualitative} and Table~\ref{tab:rebuttal-comparison}. As RFP and NeRF-SOS do not release code for multi-object segmentation, we fall back to foreground/background segmentation and use pixel classification accuracy and mIoU as the metrics.

\subsection{Ablation Study}
We perform ablations on the regularization loss and the 2D mask refinement of our method. Results are given in Table~\ref{tab:ablation} and Figure~\ref{fig:ablation-reg}. 

\vspace{-2mm}
\subsubsection{Instance Regularization Loss}
As shown in Table~\ref{tab:ablation}, the additional regularization loss $\mathcal{L}_i$ consistently improves the mIoU and PQ of our methods. The loss helps smooth fragmented segmentation regions induced by inconsistent 2D segmentation maps and close small holes on a continuous instance mask, which also suppresses noisy mask predictions floating around the objects for some scenes. Figure~\ref{fig:ablation-reg} shows the visual comparison. Regularization generally helps suppress the noise in class labels resulting from inconsistent multi-view segmentation, and partially closes the holes in segmentation caused by false negative predictions from Mask2Former.

\vspace{-2mm}
\renewcommand\width{-0.15cm}
\renewcommand\height{-0.2cm}
\newcommand{\gt}{GT}
\newcommand{\noreg}{w/o $\mathcal{L}_i$}
\newcommand{\reg}{w/ $\mathcal{L}_i$}
\begin{figure}[ht]
    \centering
    \captionsetup[subfloat]{position=below, labelformat=empty}

    
    

    \subfloat[\gt]{\includegraphics[width = 0.325\linewidth]{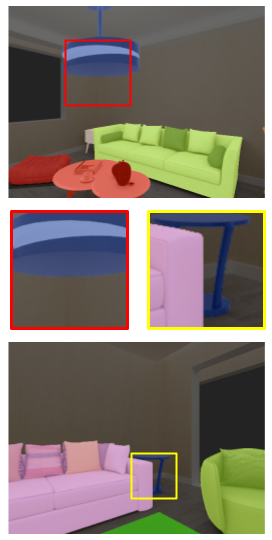}}\hspace{\width}
    \subfloat[\noreg]{\includegraphics[width = 0.325\linewidth]{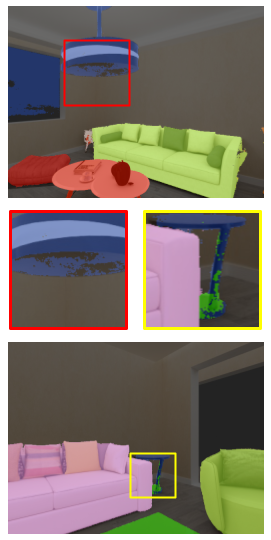}}\hspace{\width}
    \subfloat[\reg]{\includegraphics[width = 0.325\linewidth]{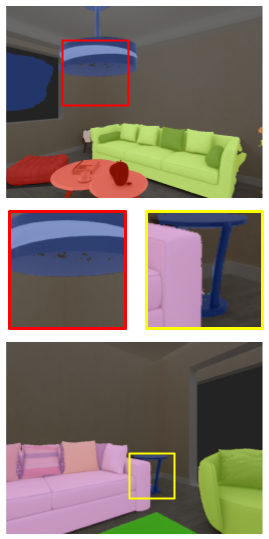}} 
    
    \caption{\textbf{Ablation on instance label regularization.} Results are taken from the \inerf before going through the refinement process.}
    \vspace{-0.2cm}
    \label{fig:ablation-reg}
\end{figure}
\vspace{-7mm}

\subsubsection{Mask Refinement}  
Using CascadePSP to refine the segmentation of the partially trained instance field helps improve the mask consistency over different frames, especially for missing predictions which leads to wrong supervision for the instance field training and cannot be easily fixed with the instance regularization. Table~\ref{tab:ablation} shows that such refinement significantly improves the mIoU and PQ of the model. For qualitative results, please refer to our Supplementary Material.

\section{Conclusion}

In this paper, we propose the {\em \inerflong} and the corresponding training approach and techniques to perform continuous 3D instance segmentation in NeRF without using ground-truth segmentation information during inference. Extensive experiments and ablation studies are performed on a synthetic indoor NeRF dataset to demonstrate the effectiveness of our method. We perform comparison to  relevant 2D segmentation methods and prior works in NeRF segmentation. As one of the first successful attempts on 3D instance segmentation in \nerf, we believe that our \inerf will contribute significantly to fundamental research on detection, segmentation in \nerf, as well as down-stream applications leveraging \nerf representation.


{\small
\bibliographystyle{ieee_fullname}
\bibliography{egbib}
}

\newcommand{\nrcnn}{NeRF-RCNN\xspace}
\setcounter{page}{0}
\pagenumbering{arabic}
\setcounter{page}{1}

\setcounter{table}{0}
\renewcommand{\thetable}{\thesection.\arabic{table}}
\renewcommand{\thefigure}{\thesection.\arabic{figure}}
\appendix


\section{NeRF 3D Instance Segmentation Dataset}
\label{sec:dataset}
Leveraging 3D-FRONT~\cite{fu20213d} and the data generating approach of~\cite{nerf-rpn}, we produce a new benchmark for instance-level 3D scene understanding curated for \nerf. 3D-FRONT is a large-scale synthetic indoor scene dataset, from which NeRF-RPN renders RGB images and layout configuration and tailors it as a benchmark for object detection task in NeRF. As shown in Table~\ref{tab:dataset-comp}, apart from multi-view images with camera poses and ground truth 3D bounding boxes, 2D ground truth instance segmentation and 3D ground truth instance masks on grids with class labels are included in our new dataset, which can be used for 3D segmentation in \nerf and other research areas. 

\begin{table}[H]
\begin{tabular}{lcc}
\hline
Dataset                            & NeRF-RPN           & Ours           \\ \hline
\# scenes                          & 152                & 1015           \\
RGB images                         & \checkmark         & \checkmark     \\ 
Camera poses                              & \checkmark         & \checkmark     \\ 
3D bounding boxes                            & \checkmark         & \checkmark     \\ 
2D inst seg GT                     &  -                 & \checkmark     \\ 
3D voxelized inst seg GT           &  -                 & \checkmark     \\ \hline
\end{tabular}

\caption{A comparison between the 3D-FRONT \nerf dataset in NeRF-RPN and ours.}
\label{tab:dataset-comp}
\end{table}

\section{\nrcnn Architecture}
In this section, we describe the architecture of NeRF-RCNN in detail. \nrcnn is a proposal-based 3D mask prediction model that imitates the architecture of Mask-RCNN~\cite{he2017mask}. The input of \nrcnn are the 3D radiance and density grid sampled from a pre-trained \nerf, and the Region of Interests (RoI) provided by NeRF-RPN~\cite{nerf-rpn}. For each RoI, we set the ground truth box with the largest intersection over union (IoU) as its regression target. 

The first part of \nrcnn is a backbone identical to~\cite{nerf-rpn} for feature extraction. The second part takes the feature of each RoI as input and predicts the 3D bounding box, classification probability and discrete 3D mask. To obtain the feature of a single proposal on a feature map, we extend RoIAlign~\cite{he2017mask} with one more dimension, making all RoI features consistent. Aligned features are fed into two heads, namely {\em box head} and {\em mask head}. {\em Box head} first flattens the inputs for fully connected layer encoding and then separates into box branch and classification branch. The box branch further regresses a RoI to a more accurate bounding box for each class, while the classification branch predicts the classification scores. We follow similar network architecture in \cite{he2017mask} by changing the 2D convolution and strided convolution layers to their corresponding 3D version. The loss function of {\em box head} consists of two parts:
\begin{equation}
      \mathcal{L}_{cls} =
    \frac{1}{|\mathcal{N}|}\sum_{i \in \mathcal{N}} \mathcal{L}_{BCE}(\boldsymbol p_i, \boldsymbol p_i^*) ,
\end{equation}
\begin{equation}
        \mathcal{L}_{reg} = \frac{1}{|\mathcal{N}_{p}|}\sum_{i\in \mathcal{N}_{p}} \sum_{k=1}^L p_{i,k}^* \mathcal{L}_{smooth}(\boldsymbol t_{i,k}, \boldsymbol t_i^*),
\end{equation}
where $\boldsymbol p_i$ is the predicted classification score vector after sigmoid, $p_{i,k}$ is the $k-$th dimension of $\boldsymbol p_i$, $\boldsymbol t_{i,k}$ is the box offsets of class $k$ , $\boldsymbol p^*_i , \boldsymbol t^*_i$ are ground-truth targets, $\mathcal{N}$ is the set of sampled RoIs, $\mathcal{N}_{p}$ is the set of positive samples, and $L$ is the number of classes including background. $\mathcal{L}_{BCE}$ and $\mathcal{L}_{smooth}$ denote the binary cross entropy(BCE) loss and the smooth L1 loss in~\cite{girshick2015fast} respectively. Note that for an RoI associated with ground truth class $c$, only the $c-$th box regression BCE loss contributes to the total loss. $\boldsymbol t_{i,k} = (t_{x,k}, t_{y,k}, t_{z,k}, t_{w,k}, t_{l,k}, t_{h,k})$ is the box head output. The relationship between $\boldsymbol t_{i,k}$ and bounding box parameters $x,y,z,w,h,l$ is defined similarly to ~\cite{nerf-rpn}:
\begin{equation}
    \begin{alignedat}{2}
    & t_{x,k} = (x_k - x_a) / w_a,\quad && t_{y,k} = (y_k - y_a) / l_a, \\
    & t_{z,k} = (z_k - z_a) / h_a, && t_{z,k} = \log(w_k / w_a), \\
    & t_{l,k} = \log(l_k / l_a), && t_{h,k} = \log(h_k / h_a), \\
\end{alignedat}
\end{equation}
where $x_k,y_k,z_k$ are the center coordinate, $w_k,l_k,h_k$ are the lengths of sides, and $x_a, y_a, z_a, w_a, l_a, h_a$ are the corresponding parameter of the RoI.

The mask head is a convolutional neural network which predicts $L$ binary masks with size $m \times m \times m$ for each RoI. $m=5$ is used for the box head, and $m=10$ for the mask head. We also apply the sigmoid function as the activation. The loss for the mask head $\mathcal{L}_M$ is defined as 
\begin{equation}
    \mathcal{L}_M = \frac{\lambda}{|\mathcal{N}_{p}|}\sum_{i\in \mathcal{N}_{p}} \sum_{k=1}^L p_{i,k}^* \mathcal{L}_{p}(\boldsymbol m_{i,k}, \boldsymbol m_i^*),
\end{equation}
where $m_i^*$ is the ground truth mask and $m_{i,k}$ is the predicted mask of class $k$. Similar to box regression, only the mask BCE loss corresponding to the ground truth label is included in $\mathcal{L}_M$.

The total loss of \inerf $\mathcal{L}$ is 
\begin{equation}
    \mathcal{L} = \mathcal{L}_{cls} + \lambda_1\mathcal{L}_{reg} + \lambda_2\mathcal{L}_{mask},
\end{equation}
where $\lambda_1, \lambda_2$ are hyper-parameters.

\section{Qualitative Results of Ablation}
We present additional visualization and qualitative comparisons to demonstrate the effectiveness of our proposed mask refinement stage. 

\begin{figure}[ht]
    \centering
    \captionsetup[subfloat]{position=below, labelformat=empty}

    \subfloat{\includegraphics[width = 0.32\linewidth]{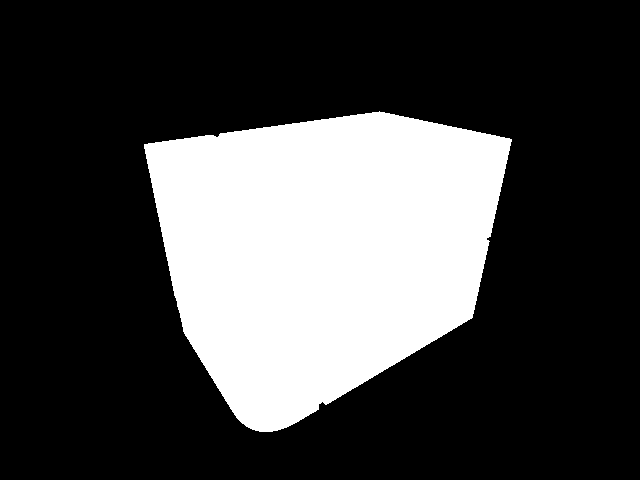}}\hspace{\width}
    \subfloat{\includegraphics[width = 0.32\linewidth]{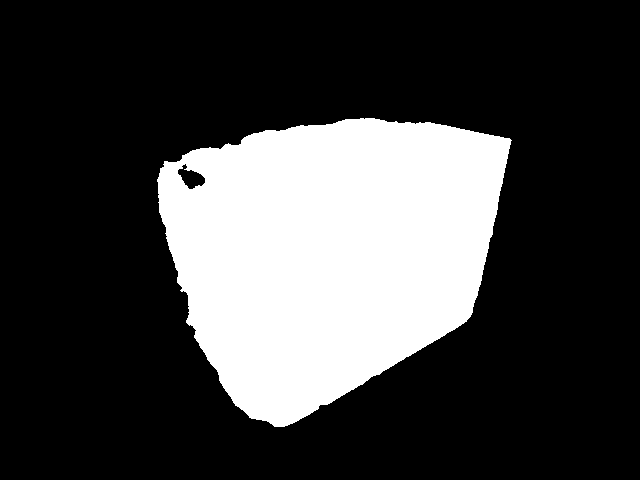}}\hspace{\width}
    \subfloat{\includegraphics[width = 0.32\linewidth]{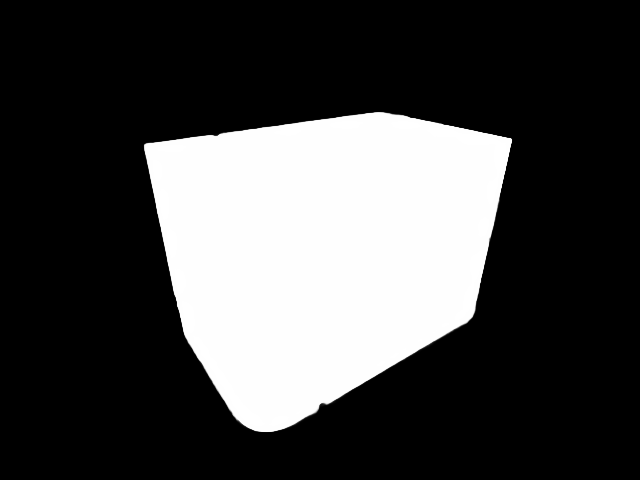}}\hspace{\width}
    
    \vspace{\height}

    \subfloat[GT]{\includegraphics[width = 0.32\linewidth]{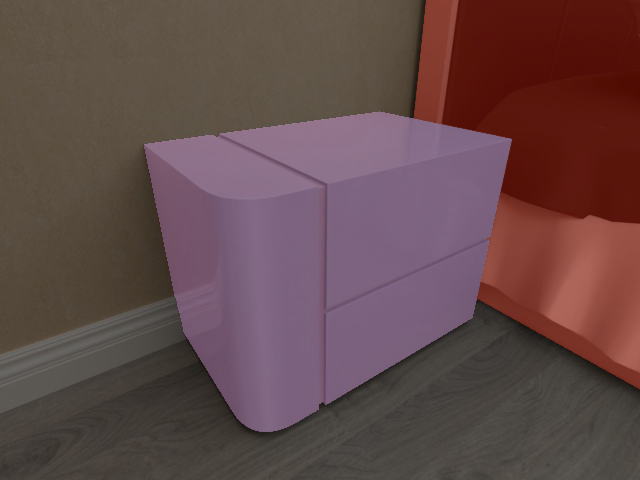}}\hspace{\width}
    \subfloat[w/o refinement]{\includegraphics[width = 0.32\linewidth]{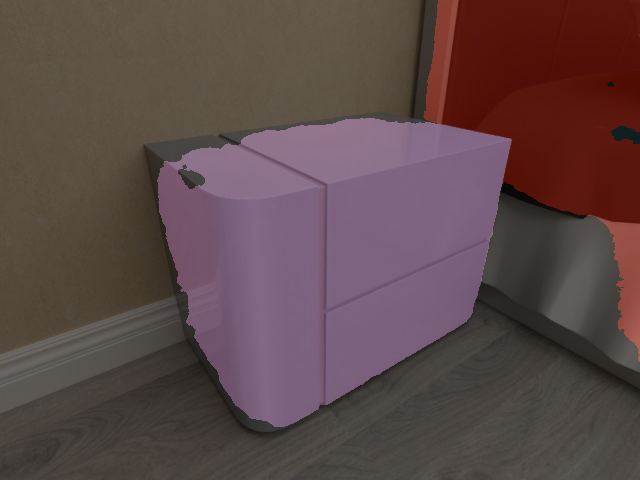}}\hspace{\width}
    \subfloat[w/ refinement]{\includegraphics[width = 0.32\linewidth]{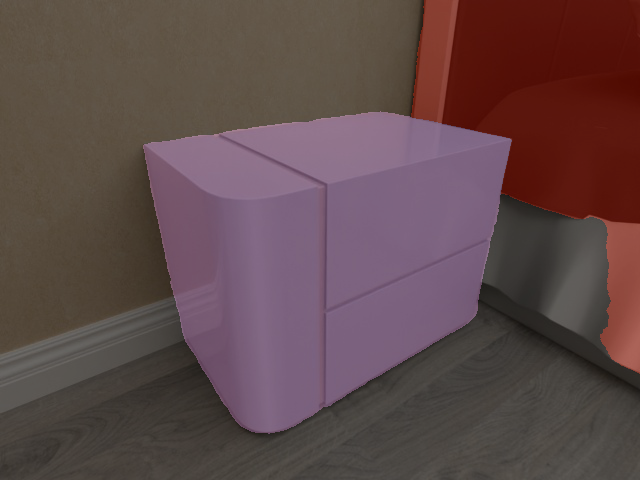}}\hspace{\width}
    
    \hspace{\width}
    \caption{\textbf{Ablation on 2D mask refinement.} The first row shows the separate mask for the nightstand, which is used as the input to CascadePSP. The separate masks after refinement are then composed into a single segmentation map to further optimize the instance field, the results of which are presented in the bottom row.}
    \vspace{0.2cm}
    \label{fig:ablation-refine}
\end{figure}

Although adding instance label regularization can help smooth the instance field, the segmentation quality of the preliminary results can still be unsatisfactory, especially on the silhouette of the objects. Besides, regularization can sometimes smooth out detailed structures in the segmentation, like thin chair legs or lamp stands. As illustrated in Figure~\ref{fig:ablation-refine}, performing 2D mask refinement using CascadePSP on the \inerf results and using it to guide further training can significantly improve the segmentation quality on the object boundaries.

\vspace{-0.1in}

\section{Additional Qualitative Comparison}
\begin{figure*}[ht]
    \centering
    \captionsetup[subfloat]{position=top, labelformat=empty}




    \subfloat[Mask2Former]{\includegraphics[width = 0.18\linewidth]{figs/3dfront_0455_00/0248.jpg}}\hspace{\resultWidth}
    \subfloat[Semantic-NeRF]{\includegraphics[width = 0.18\linewidth]{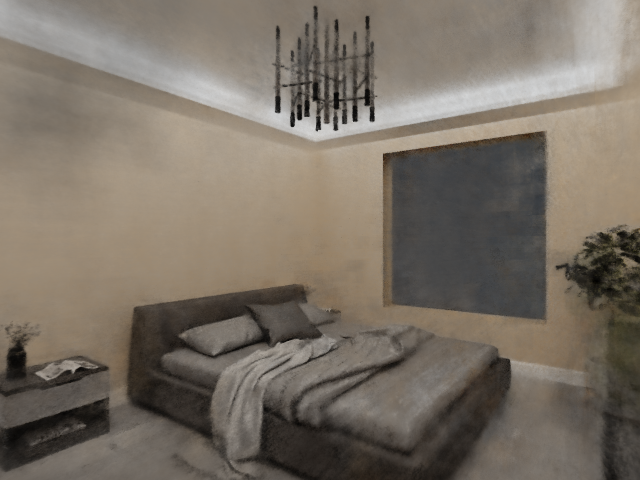}}\hspace{\resultWidth}
    \subfloat[DM-NeRF]{\includegraphics[width = 0.18\linewidth]{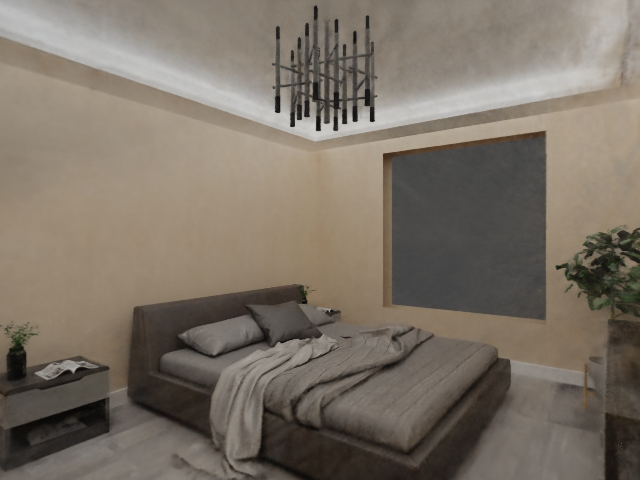}}\hspace{\resultWidth}
    \subfloat[Ours]{\includegraphics[width = 0.18\linewidth]{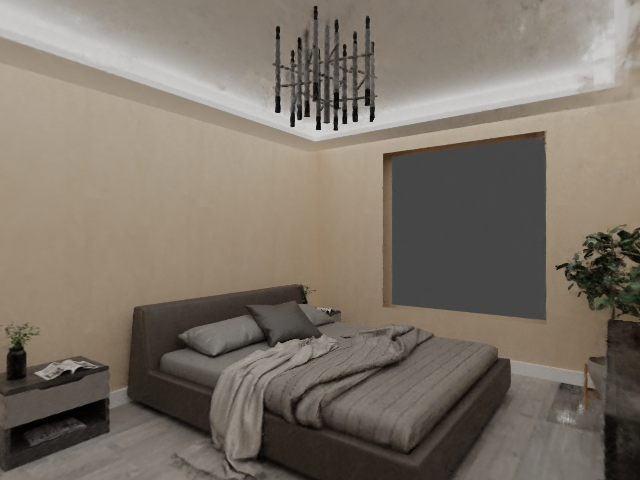}}\hspace{\resultWidth}  
    \subfloat[GT]{\includegraphics[width = 0.18\linewidth]{figs/3dfront_0455_00/0248.jpg}}

    \vspace{\resultHeight}

    \subfloat{\includegraphics[width = 0.18\linewidth]{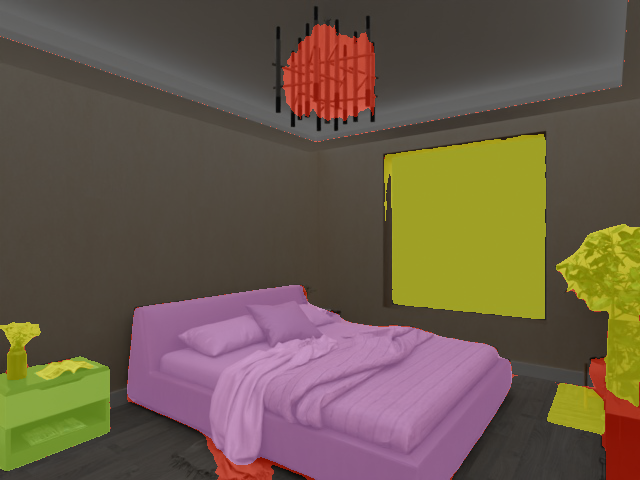}}\hspace{\resultWidth}
    \subfloat{\includegraphics[width = 0.18\linewidth]{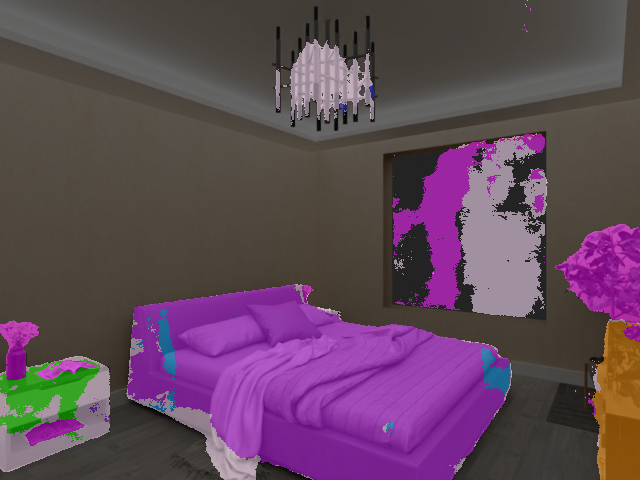}} \hspace{\resultWidth}
    \subfloat{\includegraphics[width = 0.18\linewidth]{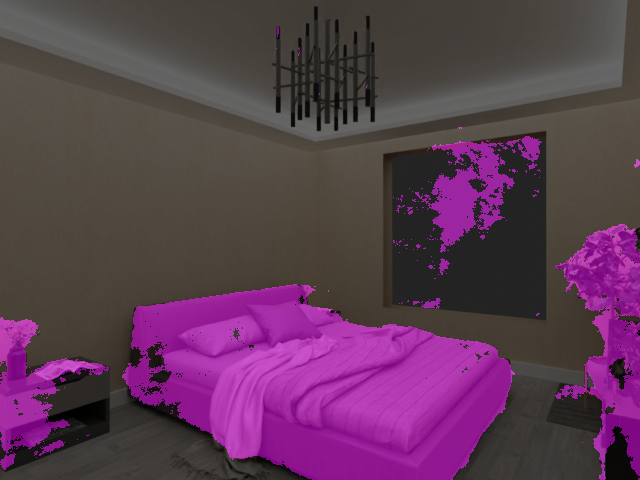}}\hspace{\resultWidth}
    \subfloat{\includegraphics[width = 0.18\linewidth]{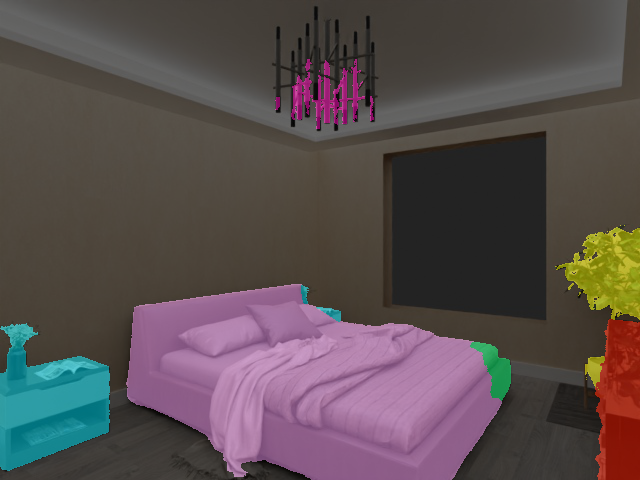}}\hspace{\resultWidth}    
    \subfloat{\includegraphics[width = 0.18\linewidth]{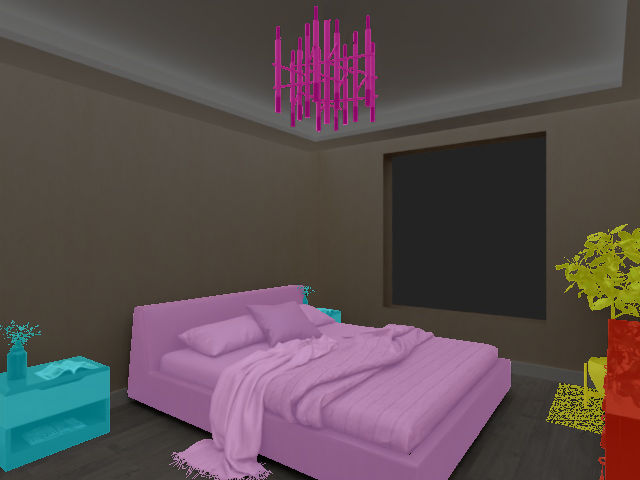}}

    \vspace{\resultHeight}

    \subfloat{\includegraphics[width = 0.18\linewidth]{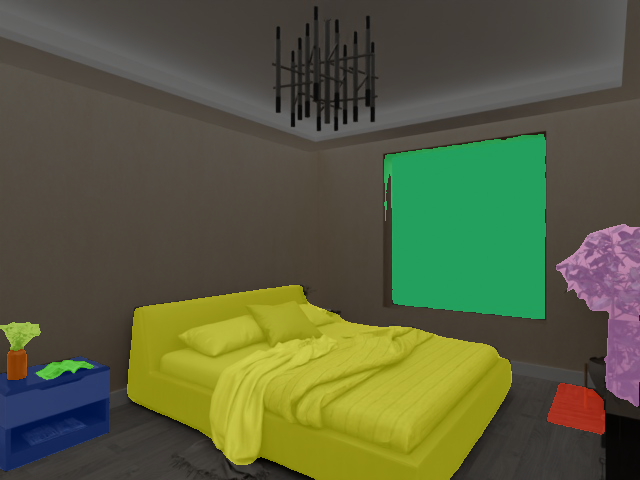}}\hspace{\resultWidth}
    \subfloat{\includegraphics[width = 0.18\linewidth]{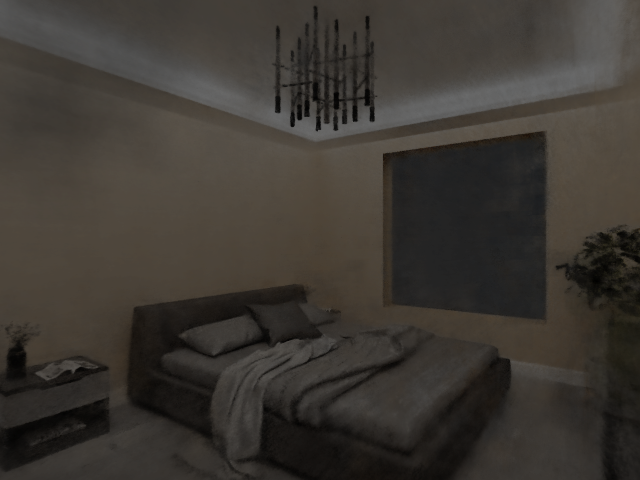}} \hspace{\resultWidth}
    \subfloat{\includegraphics[width = 0.18\linewidth]{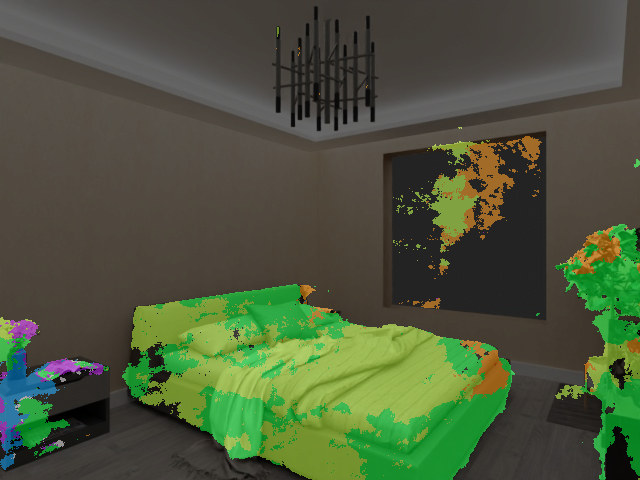}}\hspace{\resultWidth}	
    \subfloat{\includegraphics[width = 0.18\linewidth]{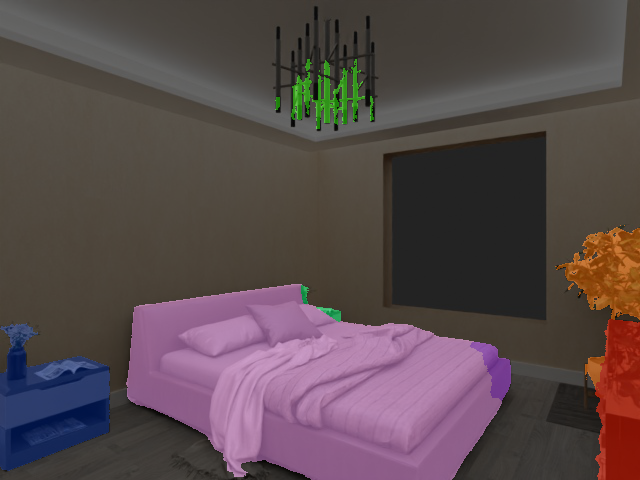}}\hspace{\resultWidth}
    \subfloat{\includegraphics[width = 0.18\linewidth]{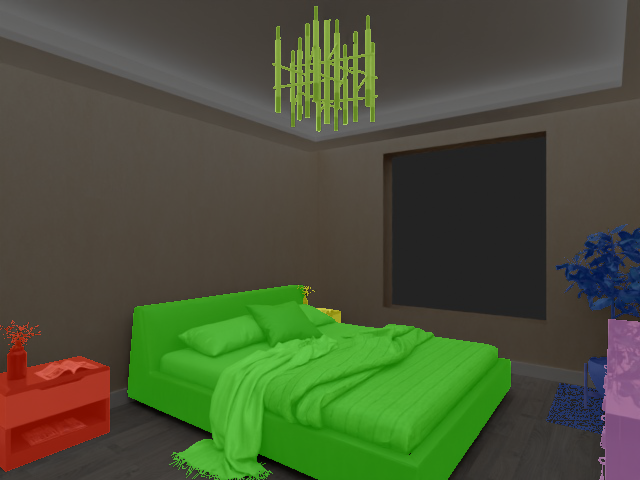}}


    \subfloat[Mask2Former]{\includegraphics[width = 0.18\linewidth]{figs/3dfront_0075_01/0201.jpg}}\hspace{\resultWidth}
    \subfloat[Semantic-NeRF]{\includegraphics[width = 0.18\linewidth]{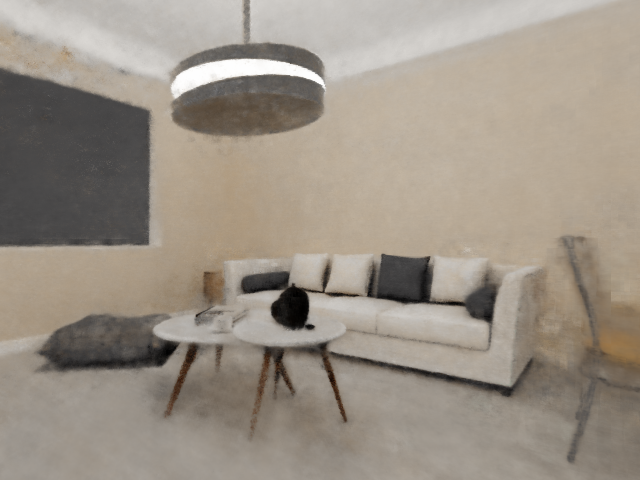}}\hspace{\resultWidth}
    \subfloat[DM-NeRF]{\includegraphics[width = 0.18\linewidth]{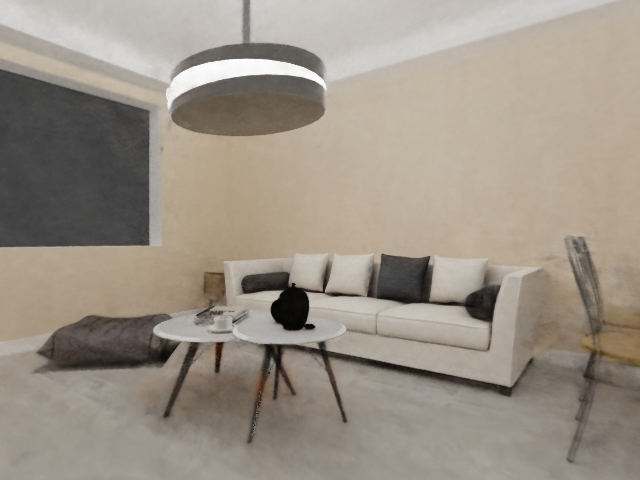}}\hspace{\resultWidth}
    \subfloat[Ours]{\includegraphics[width = 0.18\linewidth]{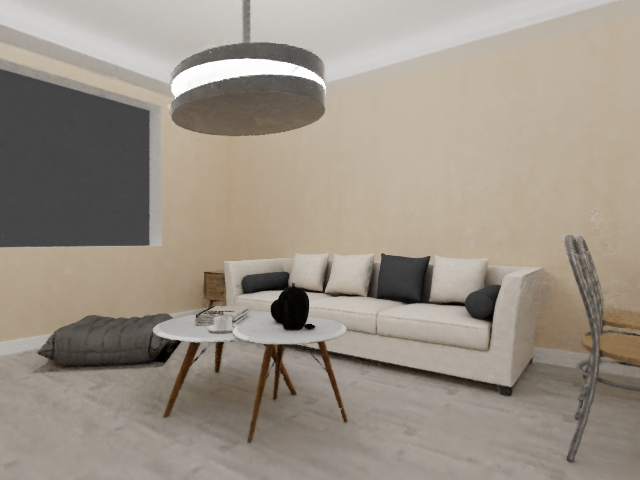}}\hspace{\resultWidth}  
    \subfloat[GT]{\includegraphics[width = 0.18\linewidth]{figs/3dfront_0075_01/0201.jpg}}

    \vspace{\resultHeight}

    \subfloat{\includegraphics[width = 0.18\linewidth]{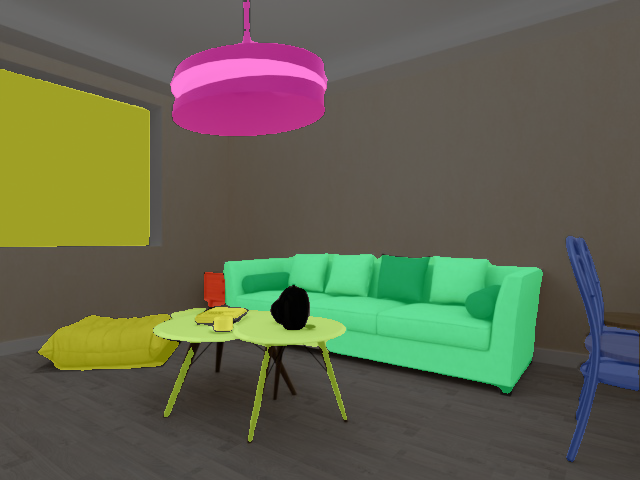}}\hspace{\resultWidth}
    \subfloat{\includegraphics[width = 0.18\linewidth]{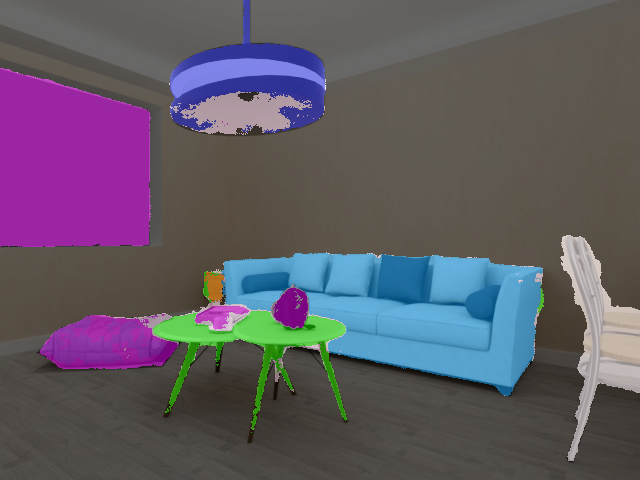}} \hspace{\resultWidth}
    \subfloat{\includegraphics[width = 0.18\linewidth]{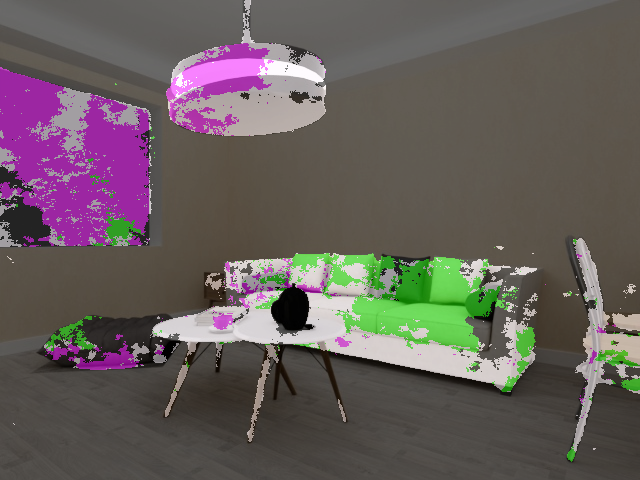}}\hspace{\resultWidth}
    \subfloat{\includegraphics[width = 0.18\linewidth]{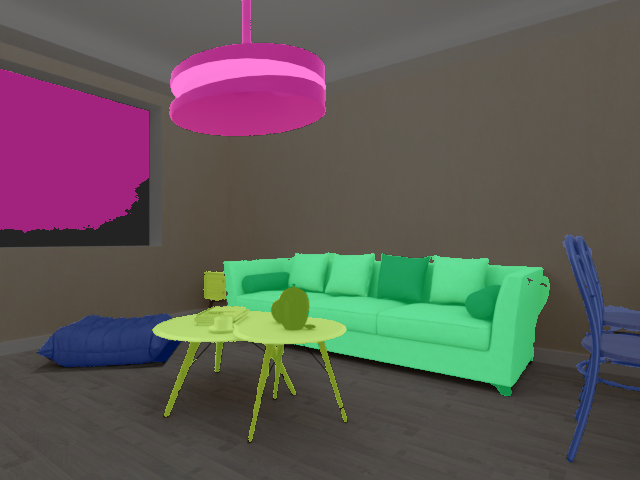}}\hspace{\resultWidth} 
    \subfloat{\includegraphics[width = 0.18\linewidth]{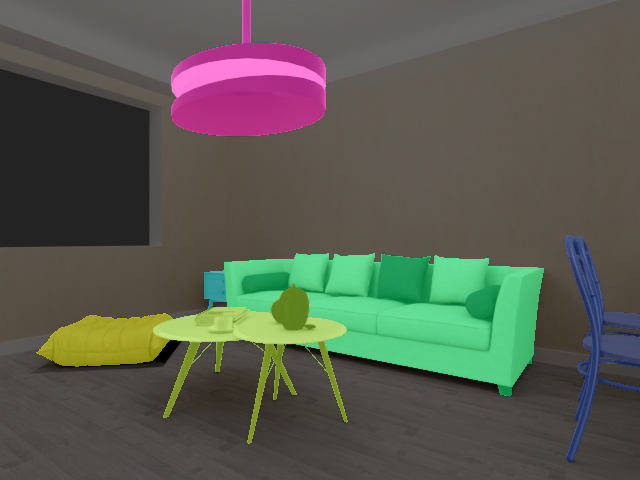}}

    \vspace{\resultHeight}
    
    \subfloat{\includegraphics[width = 0.18\linewidth]{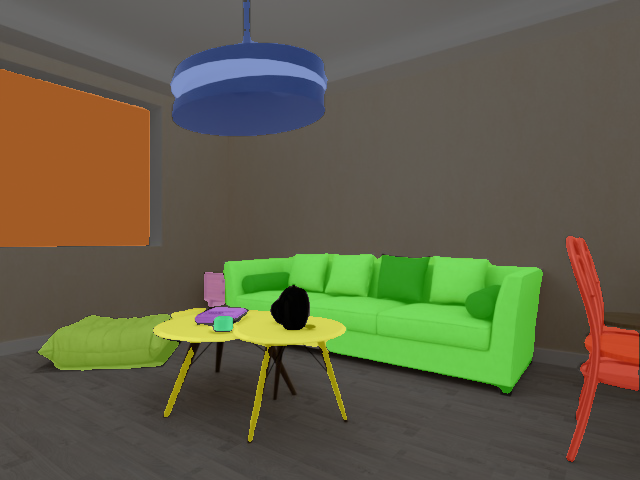}}\hspace{\resultWidth}
    \subfloat{\includegraphics[width = 0.18\linewidth]{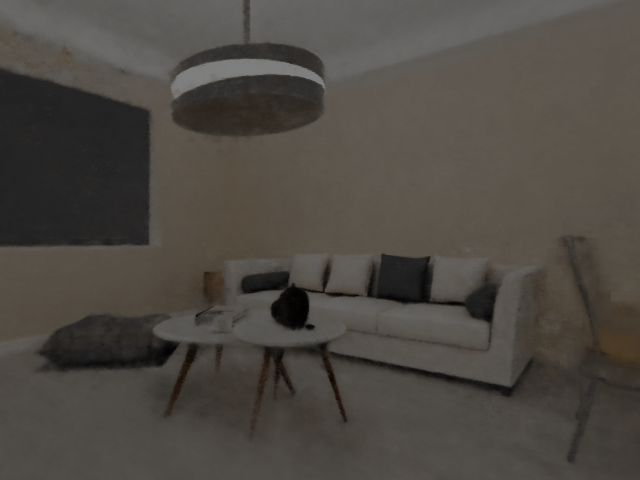}} \hspace{\resultWidth}
    \subfloat{\includegraphics[width = 0.18\linewidth]{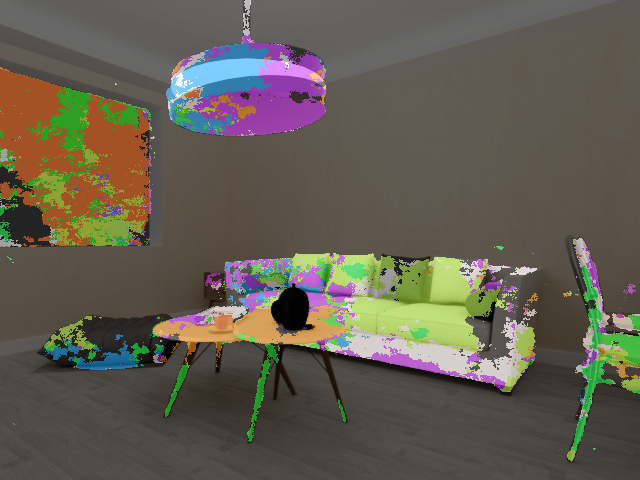}}\hspace{\resultWidth}	
    \subfloat{\includegraphics[width = 0.18\linewidth]{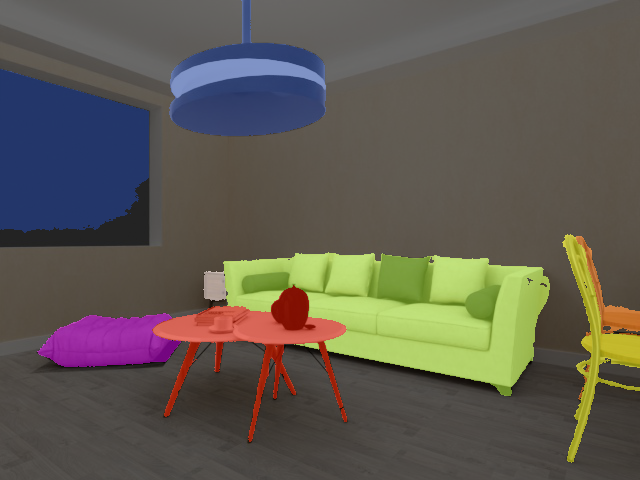}}\hspace{\resultWidth}
    \subfloat{\includegraphics[width = 0.18\linewidth]{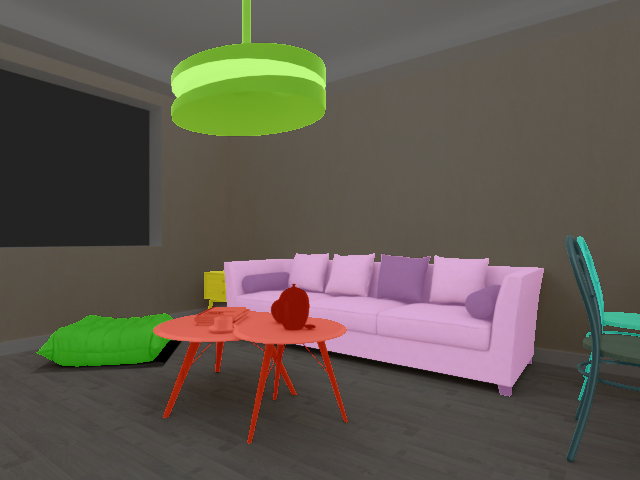}}

\caption{\textbf{Additional Comparison.} This figure illustrates the comparison between ours and other methods. For each group of comparison, rows from top to bottom are i. ground truth RGB images or the rendered RGB images from the NeRF models, ii. semantic segmentation, and iii. instance segmentation. The instance segmentation results from Semantic-NeRF are left empty as it does not produce instance-level information.}

\end{figure*}

\begin{figure*}[ht]\ContinuedFloat
    \centering
    \captionsetup[subfloat]{position=top, labelformat=empty}


    \subfloat[Mask2Former]{\includegraphics[width = 0.18\linewidth]{figs/3dfront_0261_02/0122.jpg}}\hspace{\resultWidth}
    \subfloat[Semantic-NeRF]{\includegraphics[width = 0.18\linewidth]{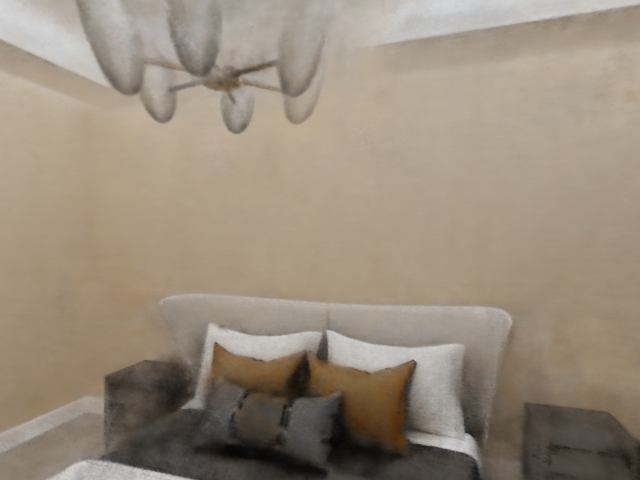}}\hspace{\resultWidth}
    \subfloat[DM-NeRF]{\includegraphics[width = 0.18\linewidth]{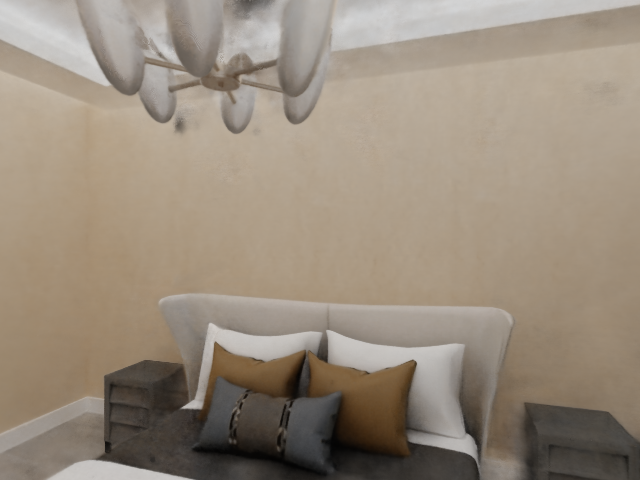}}\hspace{\resultWidth}
    \subfloat[Ours]{\includegraphics[width = 0.18\linewidth]{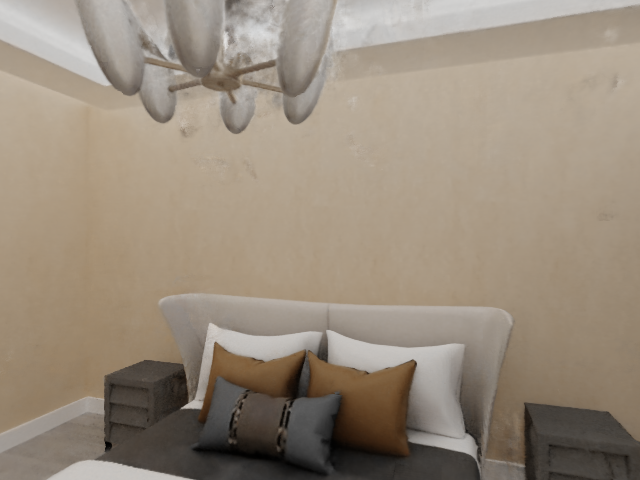}}\hspace{\resultWidth}  
    \subfloat[GT]{\includegraphics[width = 0.18\linewidth]{figs/3dfront_0261_02/0122.jpg}}

    \vspace{\resultHeight}

    \subfloat{\includegraphics[width = 0.18\linewidth]{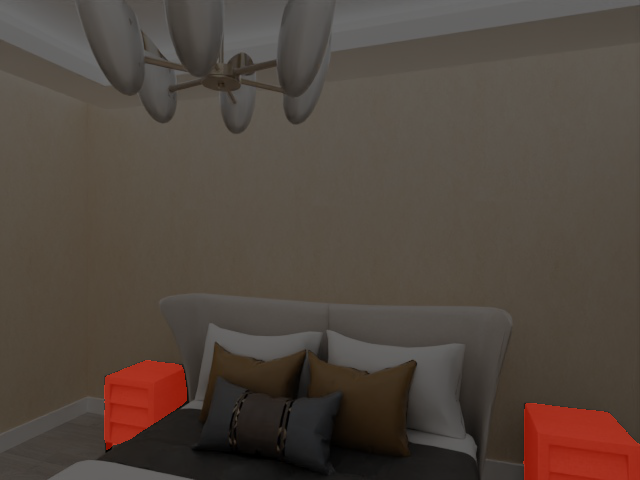}}\hspace{\resultWidth}
    \subfloat{\includegraphics[width = 0.18\linewidth]{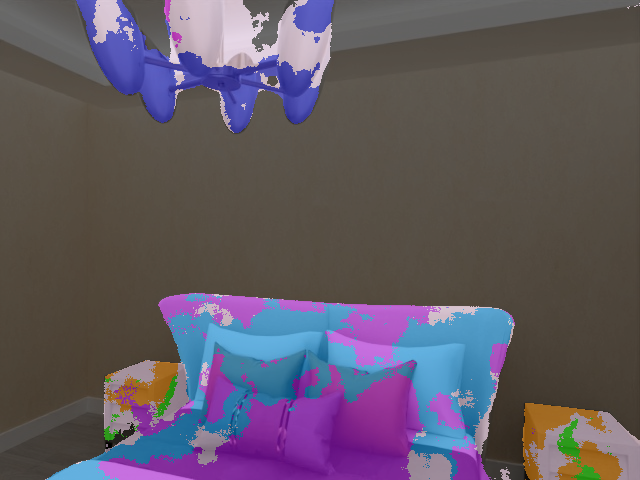}} \hspace{\resultWidth}
    \subfloat{\includegraphics[width = 0.18\linewidth]{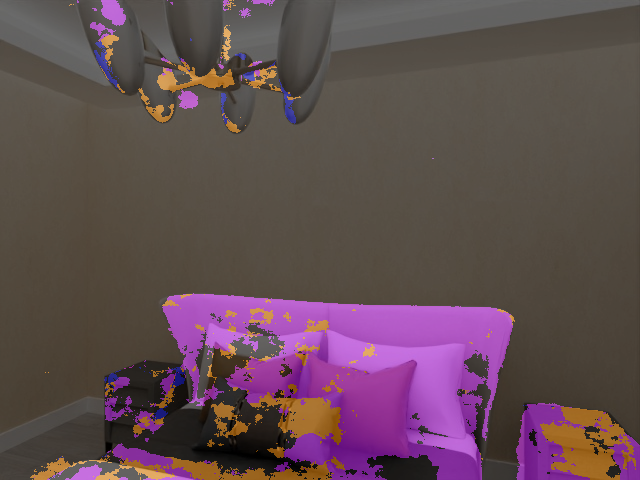}}\hspace{\resultWidth}
    \subfloat{\includegraphics[width = 0.18\linewidth]{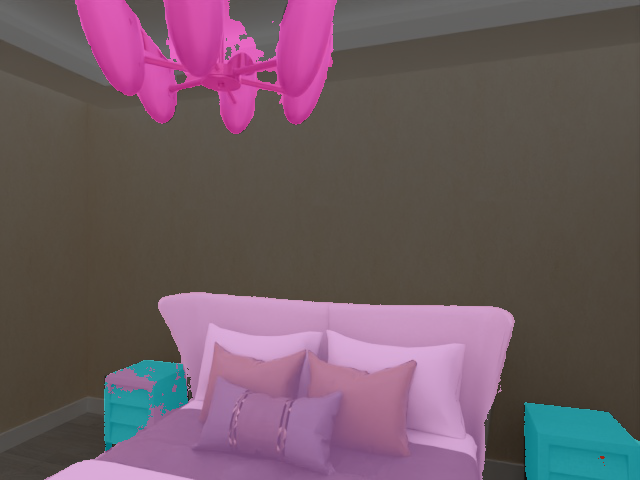}}\hspace{\resultWidth} 
    \subfloat{\includegraphics[width = 0.18\linewidth]{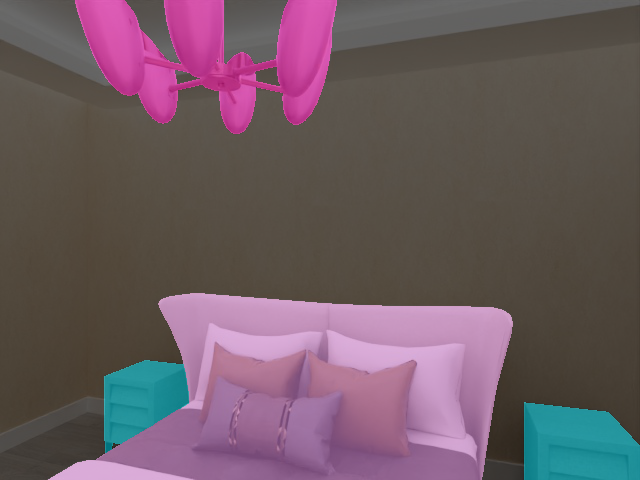}}

    \vspace{\resultHeight}
    
    \subfloat{\includegraphics[width = 0.18\linewidth]{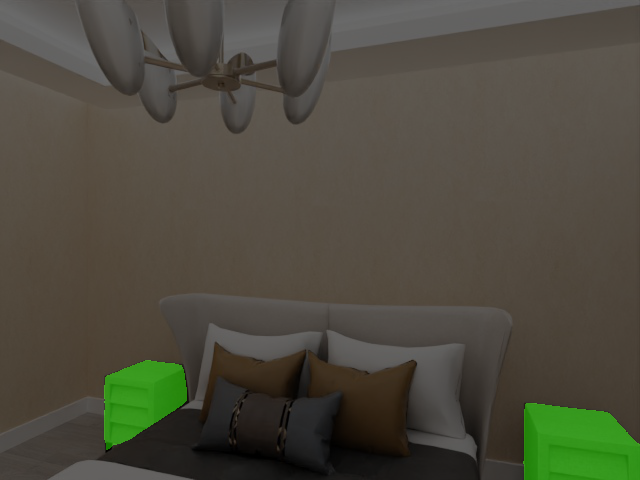}}\hspace{\resultWidth}
    \subfloat{\includegraphics[width = 0.18\linewidth]{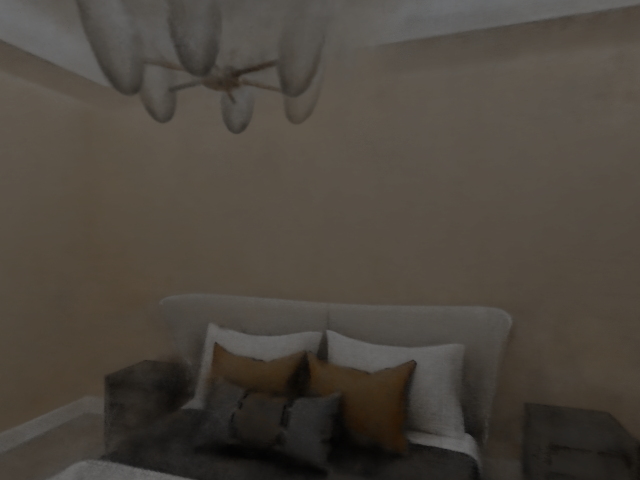}} \hspace{\resultWidth}
    \subfloat{\includegraphics[width = 0.18\linewidth]{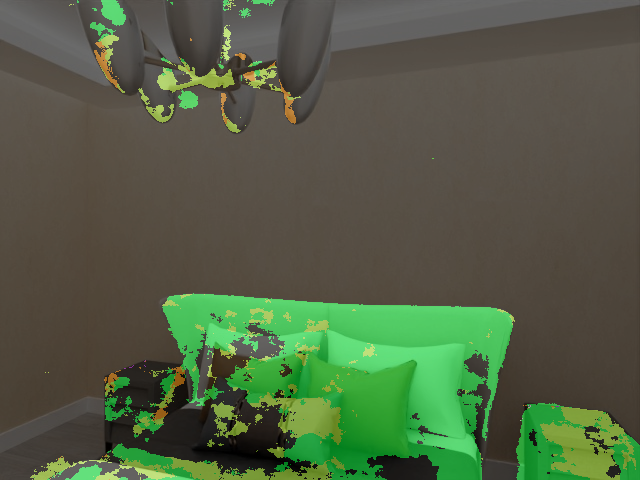}}\hspace{\resultWidth}	
    \subfloat{\includegraphics[width = 0.18\linewidth]{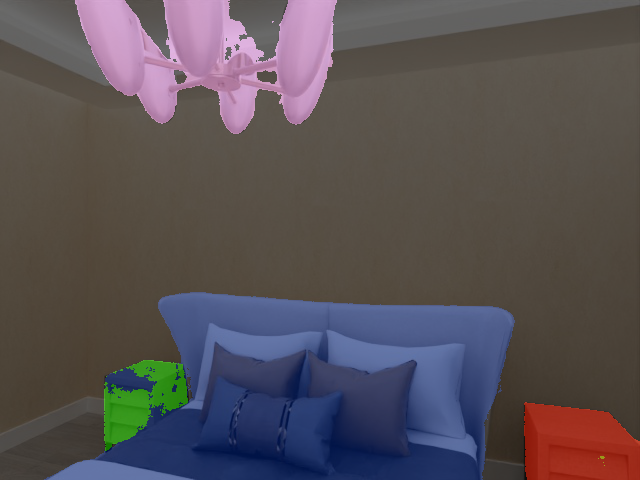}}\hspace{\resultWidth}
    \subfloat{\includegraphics[width = 0.18\linewidth]{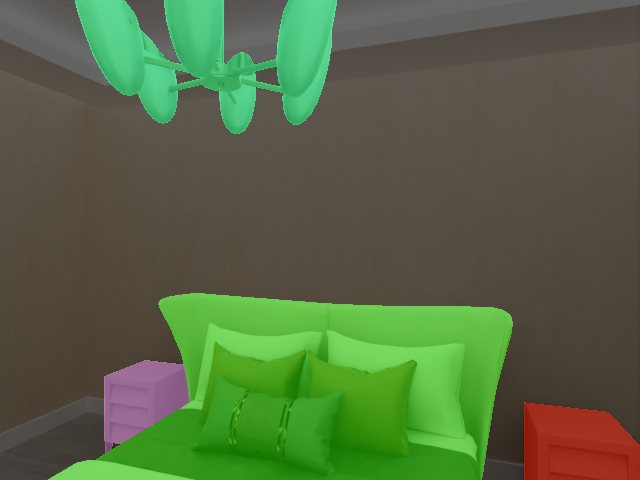}}

\caption{\textbf{Additional Comparison (cont.)} This figure illustrates the comparison between ours and other methods. For each group of comparison, rows from top to bottom are i. ground truth RGB images or the rendered RGB images from the NeRF models, ii. semantic segmentation, and iii. instance segmentation. The instance segmentation results from Semantic-NeRF are left empty as it does not produce instance-level information.}

    \label{fig:comparison-extra}
\end{figure*}

We demonstrate extra qualitative comparisons between our method and other related methods as mentioned in the main paper. The results are given in Figure~\ref{fig:comparison-extra}. Please watch the video at \url{https://www.youtube.com/watch?v=wW9Bme73coI} for more qualitative comparison results.

\end{document}